\documentclass[10pt,twocolumn,letterpaper, dvipsnames]{article}

\usepackage{iccv}
\usepackage{times}
\usepackage{epsfig}
\usepackage{graphicx}
\usepackage{amsmath}
\usepackage{amssymb}

\usepackage{cite}
\usepackage{csvsimple}
\usepackage{booktabs}
\usepackage{pgfplots}
\usepackage{subcaption}
\usetikzlibrary{matrix}
\usetikzlibrary{fillbetween}
\usetikzlibrary{intersections}
\usepgfplotslibrary{groupplots}
\usepgfplotslibrary{fillbetween}
\pgfplotsset{compat = newest}
\usepackage{array}
\usepackage{colortbl}
\usepackage{multirow}
\usepackage{xfp}
\usepackage{hhline}
\usepackage{siunitx}
\usepackage{pifont}

\usepackage{algorithm}
\usepackage{algpseudocode}

\DeclareMathOperator*{\argmax}{arg\,max}

\usepackage[pagebackref=true,breaklinks=true,letterpaper=true,colorlinks,bookmarks=false]{hyperref}

\iccvfinalcopy 

\def\iccvPaperID{7} 

\ificcvfinal\fi

\begin{document}

\title{HyperSparse Neural Networks: Shifting Exploration to Exploitation through Adaptive Regularization}

\author{Patrick Glandorf$^{*}$, Timo Kaiser$^{*}$, Bodo Rosenhahn\\
Institute for Information Processing (tnt)\\
L3S - Leibniz Universität Hannover, Germany\\
{\tt\small \{glandorf, kaiser, rosenhahn\}@tnt.uni-hannover.de}
}

\maketitle
\ificcvfinal\thispagestyle{empty}\fi

\def\thefootnote{*}\footnotetext{These authors contributed equally to this work}\def\thefootnote{\arabic{footnote}}
\begin{abstract}
Sparse neural networks are a key factor in developing resource-efficient machine learning applications. We propose the novel and powerful sparse learning method \textit{Adaptive Regularized Training} (ART) to compress dense into sparse networks. Instead of the commonly used binary mask during training to reduce the number of model weights, we inherently shrink weights close to zero in an iterative manner with increasing weight regularization. Our method compresses the pre-trained model ``knowledge'' into the weights of highest magnitude. Therefore, we introduce a novel regularization loss named \textit{HyperSparse} that exploits the highest weights while conserving the ability of weight exploration. Extensive experiments on CIFAR and TinyImageNet show that our method leads to notable performance gains compared to other sparsification methods, especially in extremely high sparsity regimes up to $99.8\%$ model sparsity. Additional investigations provide new insights into the patterns that are encoded in weights with high magnitudes.\footnote{Code available at \href{https://github.com/GreenAutoML4FAS/HyperSparse}{https://github.com/GreenAutoML4FAS/HyperSparse}}
\end{abstract}
\vspace{-20pt}
\begin{figure}[t]
\centering
\resizebox{\columnwidth}{!}{ 
\newcommand{\fileSortWeight}{ablation_results/sortet_weights_by_value/sepGrad_RegTerm=HyperSparse_dataset=cifar100_model=resnet32_pruneRate=0.9.txt}

\begin{tikzpicture}
\begin{groupplot}[group style={group size= 1 by 2},
                    height = 0.2\linewidth,width=\columnwidth]
    
    \pgfplotsset{/pgfplots/group/.cd, vertical sep=0.1cm}
                
    \nextgroupplot[title=Weight Magnitudes and Gradients,
    width = 1\columnwidth, height = 0.5\linewidth,
    ylabel=$|w_i|$,
    xmin=0.65, xmax=1., 
    xticklabels=\empty,
    xtick distance = 0.1, ytick distance = 0.2,
    minor tick num = 1,
    xmajorgrids=true, ymajorgrids=true,
    xminorgrids=true, yminorgrids=true,
    major grid style = {lightgray},
    minor grid style = {lightgray!25},
    legend style = {at={(0.05,0.93)}, anchor=north west, legend columns=2}
    ]
    \addlegendimage{empty legend}
    \addlegendimage{empty legend}
    \addplot[orange] table [x=idx,y=w_060,col sep=comma] {\fileSortWeight};
    \addplot[magenta] table [x=idx,y=w_070,col sep=comma] {\fileSortWeight};
    \addplot[cyan] table [x=idx,y=w_080,col sep=comma] {\fileSortWeight};
    \addplot[blue] table [x=idx,y=w_090,col sep=comma] {\fileSortWeight};
    \draw [dashed, thick, black] (0.9,-1) -- (0.9,10);
    \legend{\hspace{-.6cm}Epoch, {\ }, ~~0, 20, 30, 40}
    
    \nextgroupplot[
    scaled ticks=false, log ticks with fixed point, tick label style={/pgf/number format/fixed},
    width = 1\columnwidth, height = 0.5\linewidth,
    xlabel=Relative Weight Index $i$, ylabel=\shortstack{ $\bigg|\frac{d\mathcal{L}_\text{HS}}{d w_i}\bigg|$},
    xmin=0.65, xmax=1., 
    ymin=0., ymax=2.5, 
    xtick distance = 0.1, ytick distance = 1,
    minor tick num = 1,    
    xmajorgrids=true, ymajorgrids=true,
    xminorgrids=true, 
    yminorgrids=true,
    major grid style = {lightgray},
    minor grid style = {lightgray!25},
    ]
    
    \addplot[orange] table [x=idx,y=g_reg_060,col sep=comma] {\fileSortWeight};
    \addplot[magenta] table [x=idx,y=g_reg_080,col sep=comma] {\fileSortWeight};
    \addplot[cyan] table [x=idx,y=g_reg_090,col sep=comma] {\fileSortWeight};
    \addplot[blue] table [x=idx,y=g_reg_100,col sep=comma] {\fileSortWeight};

    \draw [dashed, thick, black] (0.9,-1) -- (0.9,100);
    
    \draw [very thick, OliveGreen] (0.0,1.0) -- (1.0,1.0);
    \node[below,OliveGreen] at (0.68,1.0) {$\mathcal{L}_1$};
    
\end{groupplot}

\end{tikzpicture}}
\vspace{-20pt}
\caption{Magnitude of weights with their corresponding gradients at different epochs derived from our \textit{HyperSparse} loss sorted by the weight magnitude. The weights and gradients belong to a ResNet-32 trained on Cifar-100, where the desired pruning rate is $\kappa=90\%$. The smallest weight $w_\kappa$ that remains after pruning is marked by a dashed line. Note that we added the gradient for the $\mathcal{L}_1$ loss in green.}
\vspace{-10pt}
\label{fig:gradSortetWeightMag}
\end{figure}
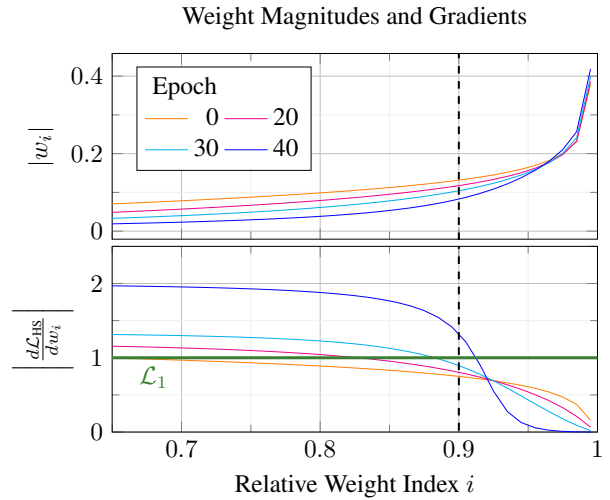

\section{Introduction}

Recent years have shown tremendous progress in the field of machine learning based on the use of neural networks (NN). Alongside the increasing accuracy in nearly all tasks, also the computational complexity of NNs increased, \eg, for Transformers~\cite{dosovitskiy2021an,cong2023reltr} or Large Language Models~\cite{NEURIPS2020_1457c0d6}. The complexity causes high energy costs, limits the applicability for cost efficient systems~\cite{NIPS2015_ae0eb3ee}, and is counterproductive for the sake of fairness and trustworthiness due to dwindling interpretability~\cite{norrenbrock2022take}.   

Facing these issues, recent years have also led to a growing community in the field of sparse NNs~\cite{hoefler2021sparsity}. The goal is to find small subgraphs (\textit{a.k.a} sparse NNs) in well performing NNs that have similar or comparable capabilities regarding the main tasks while being significantly less complex and therefore cheaper and potentially better interpretable.      
Standard methods usually create sparse NNs by obtaining a binary mask that limits the number of used weights in a NN~\cite{lee2018snip, wang2019picking, NEURIPS2020_eae27d77}. 
The most prominent method is \textit{Iterative Magnitude Pruning} (\textit{IMP})~\cite{pmlr-v119-frankle20a} that is based on the \textit{Lottery Ticket Hypothesis} (\textit{LTH})~\cite{frankle2018lottery}. Assuming that important weights have high magnitudes after training, it trains a dense NN and removes an amount of elements from the mask that correspond to the lowest weights. Afterward, the sparse NN is reinitialized and retrained from scratch. The process is iterated until a sparsity level is reached. 

The assumption of magnitude pruning that highest weights in dense NNs encode most important decision rules for a diverse set of classes is problematic, because it is not guaranteed. 
Removed weights that are potentially useful to the prediction can no longer be reactivated during fine-tuning.
In the worst case, a ``layer collapse'' can prohibit a useful forward propagation~\cite{NEURIPS2020_46a4378f}. The lack of exploration ability still persists in the more accurate but resource consuming iterative \textit{IMP} approach.

Reviving the key ideas of \textit{Han} \etal~\cite{NIPS2015_ae0eb3ee} and \textit{Narang} \etal~\cite{narang2017exploring} (comparable to~\cite{DBLP:journals/corr/MolchanovTKAK16}), we introduce a lightweight and powerful method called \textit{Adaptive Regularized Training} (\textit{ART}) to obtain highly sparse NNs, which implicitly ``removes'' weights with increasing regularization until a desired sparsity level is reached. \textit{ART} 
strongly regularizes the weights before magnitude pruning. First, a dense NN is pre-trained until convergence. In the second stage, the NN is trained with an increasing and weight decaying regularization until the hypothetical magnitude pruned NN performs on par with the dense counterpart. Lastly, we apply magnitude pruning and fine-tune the NN without regularization.
Avoiding binary masks in the second stage allows exploration and regularization forces the exploitation of weights that remain in the sparse NN. We introduce the new regularization approach \textit{HyperSparse} for the second stage that overcomes static regularization like \textit{Lasso}~\cite{Tibshirani1996RegressionSA} or \textit{Weight Decay}~\cite{https://doi.org/10.48550/arxiv.1810.12281} and adapts to the weight magnitude by penalizing small weights. \textit{HyperSparse} balances the exploration/exploitation tradeoff and thus increases the accuracy while leading to faster convergence in the second stage. 
The combination of our regularization schedule and \textit{HyperSparse} improves the classification accuracy and optimization time significantly, especially in high sparsity regimes with up to $99.8\%$ zero weights. We evaluate our method on CIFAR-10/100~\cite{krizhevsky2009learning} and TinyImageNet~\cite{deng2009imagenet} with ResNet-32~\cite{He_2016_CVPR} and VGG-19~\cite{simonyan2014very}.

Moreover, we analyze the gradient and weight distribution during regularized training, showing that \textit{HyperSparse} leads to faster convergence to sparse NNs. The experiments also shows that the claim of~\cite{NEURIPS2020_eae27d77}, that optimal sparse NNs can be obtained via simple weight distribution heuristics, does not hold in general. Finally, we analyze the process of compressing dense NNs into sparse NNs and show that the highest weights in NNs do not encode decision rules for a diverse set of classes with equal priority.  

\textbf{In summary}, this paper
\begin{itemize}
\vspace{-5pt}
    \item introduces \textit{HyperSparse}, a superior adaptive regularization loss that implicitly promotes configurable network sparsity by balancing the exploration and exploitation tradeoff.\vspace{-7pt}
    \item introduces the novel framework \textit{ART} to obtain sparse networks using regularization with increasing leverage, which improves the optimization time and classification accuracy of sparse neural networks, especially in high sparsity regimes.\vspace{-7pt}
    \item analyzes the continuous process of compressing patterns from dense to sparse neural networks.
\end{itemize}
\section{Related Work}

\paragraph{Sparse Learning} methods that find binary masks to remove a predefined amount of weights can be categorized as static or dynamic (\eg, in~\cite{chen2022sparsity,pmlr-v162-jaiswal22a, hoefler2021sparsity}). According to~\cite{chen2022sparsity}, in dynamic sparse training \textit{``[...] removed elements \text{[from masks]} have chances to be grown back if they potentially benefit to predictions''} whereas static training incorporates fixed masks. 

Static methods are usually based on \textit{Frankle} \etal~\cite{frankle2018lottery}, who introduce \textit{LTH} and show that well performing sparse NNs in random initialised NNs can be found after dense training via magnitude pruning. 
The magnitude pruning method is improved by \textit{IMP}~\cite{pmlr-v119-frankle20a} that iterates the process. Replacing the time consuming training procedure, methods like \textit{SNIP}~\cite{lee2018snip} or \textit{GraSP}~\cite{wang2019picking} find sparse NNs in random initialized dense NNs using a single network prediction and its gradients. To also address the risk of layer collapse during pruning, \textit{SynFlow}~\cite{NEURIPS2020_46a4378f} additionally conserves the total flow in the network. Contrary to the latter works, \textit{Su} \etal.~\cite{NEURIPS2020_eae27d77} claim that appropriate sparse NNs do not depend on data or weight initialization and provide a general heuristic for the distribution of weights. 

Different from static methods, dynamic methods prune and re-activate zero elements in the binary mask. 
The weights that are reactivated can be selected randomly~\cite{mocanu2018scalable} or determined by the gradients~\cite{chen2022sparsity,NEURIPS2021_a61f27ab,pmlr-v119-evci20a}.
For example, \textit{RigL}~\cite{pmlr-v119-evci20a} iteratively prunes weights with low magnitude and therefore reactivates weights with highest gradient.
Also, modern dynamic methods utilize continuous masks. For example, \textit{Tai} \etal~\cite{tai2022spartan} relax the \textit{IMP} framework by introducing a parameterized softmask to obtain a weighted average between \textit{IMP} and \textit{Top-KAST}~\cite{NEURIPS2020_ee76626e}. Similar,~\cite{NEURIPS2020_83004190,louizos2018learning} relaxes the binary mask and optimizes its $\mathcal{L}_0$-norm. Another way is to inherently prune the model, \eg, by reducing the gradients of weights with small magnitude~\cite{NEURIPS2021_f1e709e6}. 
Compared to static methods, \textit{Liu} \etal~\cite{pmlr-v139-liu21y,10.1007/978-3-030-67664-3_17} show that dynamic sparse training methods overcomes most static methods by allowing weight exploration.

Another property to distinguish modern sparse learning methods is the complexity during mask generation, \eg., as done by \textit{Schwarz} \etal~\cite{NEURIPS2021_f1e709e6}. The more resource efficient \textit{sparse$\rightarrow$sparse} methods sustain sparse NNs during training~\cite{lee2018snip, wang2019picking,NEURIPS2020_46a4378f, NEURIPS2020_eae27d77, NEURIPS2021_f1e709e6,pmlr-v119-evci20a,chen2022sparsity}, whereas \textit{dense$\rightarrow$sparse} methods utilize all parameters before finding the final mask~\cite{frankle2018lottery,pmlr-v119-frankle20a, tai2022spartan,NEURIPS2020_ee76626e,NEURIPS2020_83004190,louizos2018learning}.   

However, as explained later, our approach belongs to \textit{dense$\rightarrow$sparse} methods that inherently reduce the model complexity without masking before magnitude pruning to obtain a static sparse mask for fine-tuning. We want to mention the primary works of \textit{Han} \etal~\cite{NIPS2015_ae0eb3ee}, \textit{Narang} \etal~\cite{narang2017exploring} and \textit{Molchanov} \etal~\cite{DBLP:journals/corr/MolchanovTKAK16} whose combination is a role model for us. \textit{Han} \etal use $\mathcal{L}_1$ and $\mathcal{L}_2$ regularization to reduce the number of non-zero elements during training. Their early framework uses regularization without bells and whistles and has no ability to control the sparsity level. \textit{Narang} \etal and \textit{Molchanov} \etal remove weights in fine-grained portions with an increasing removal-threshold, but do not incorporate weight exploration.

\paragraph{Interpretability and Understanding} of machine learning is closely related to sparse learning and is also addressed in this paper. There is an increasing number of works in recent years that utilize sparse learning for other benefits, for example, to find interpretable correlations between feature- and image-space~\cite{norrenbrock2022take} or
to visualize inter-class ambiguities~\cite{Kaiser_2023_CVPR}. The work of \textit{Paul} \etal~\cite{paul2022lottery} gives details about the early learning stage which is crucial, \eg, to determine memorization of label noise~\cite{https://doi.org/10.48550/arxiv.2211.11355}.  They show that most data is not necessary to obtain suitable subnetworks. The general relationship between \textit{LTH} and generalization is investigated in~\cite{sakamoto2022analyzing}. \textit{Varma} \etal~\cite{t2022sparse} show that sparse NNs are better suited in data limited and noisy regimes. On the other hand \textit{Hooker} \etal~\cite{https://doi.org/10.48550/arxiv.1911.05248} show that sparse NNs have non-trivial impact on ethical bias by investigating which samples are ``forgotten'' first during network compression. The underlying research question of the latter work is altered to \textit{``Which samples are compressed first?''} and discussed in this paper.

\section{Method}

Sparsification aims to reduce the number of non-zero weights in a NN.
To address this problem, we use a certain schedule for regularization such that small weights converge to zero and our model implicitly becomes sparse.
In Sec.~\ref{sec:preliminaries}, we formally define the sparsification problem.
Then, we present \textit{Adaptive Regularized Training} (ART) in Sec.~\ref{sec:regularized_training}, which iteratively increases the leverage of regularization to maximize the number of close-to-zero weights.
Moreover, we introduce our regularization loss \textit{HyperSparse} in Sec.~\ref{sec:SparTan} that is integrated in \textit{ART}. 
It simultaneously allows the exploration of new topologies while exploiting weights of the final sparse subnetwork.

\subsection{Preliminaries}
\label{sec:preliminaries}

We consider a NN $f(W,x)$ with topology $f$ and weights $W$ that is trained to classify images from a dataset \hbox{$S=\{(x_n, y_n)\}^N_{n=1}$}, where $y_n$ is the ground truth class to an image sample $x_n$. 
The training is structured in epochs, which are iterative optimizations of the weights $W=\{w_1,\ldots, w_D\}$ over all samples in $S$ 
to minimize the loss objective $\mathcal{L}$. 
The obtained weights after epoch $e$ are denoted as $W_e$, with $W_0$ denoting the weights before optimization.
Furthermore, the classification accuracy of a NN is measured by a rating function $\psi(W)$.

The goal in sparsification is to reduce the cardinality of $W$ by removing a pre-defined ratio of weights $\kappa$, while maximizing $\psi(W)$.
The network is pruned by the Hadamard product $m \odot W$ of a binary mask $m \in [0,1]^D$ and the model-weights $W$.
The mask is usually created by applying magnitude pruning $m=\nu(W)$\cite{frankle2018lottery, pmlr-v119-frankle20a, chen2022sparsity, NEURIPS2021_a61f27ab}, which is a technique that sets the $\kappa$-lowest weights to zero.

\subsection{Adaptive Regularized Training (ART)}
\label{sec:regularized_training}

\begin{algorithm}[b]
\caption{Adaptive Regularized Training (ART)}

\textbf{Parameter:} 
Pre-trained weights~$W_\text{pre}$,
initial rate~$\lambda_\text{init}$, 
rating function~$\psi(W)$, 
magnitude pruning~$\nu(W)$,
increasing factor~$\eta > 1$,
classification loss~$\mathcal{L}_\text{class}$,
regularization loss~$\mathcal{L}_\text{reg}$,
training data $S$,
optimizer~$\text{SGD}(W,\mathcal{L},S)$

\textbf{Result:} 
Best weights for fine-tuning~$W_\text{best}$
\begin{algorithmic}[1]
    \State $W_0, W_\text{best} \gets W_\text{pre}$ 
    \State $e \gets 0$
    \While{$\psi(\nu(W_\text{best}) \odot W_\text{best}) < \psi(W_e)$} 
        \State $ W_{e+1} \gets \text{SGD}(W_e,\mathcal{L}_\text{class} + \lambda_\text{init} \cdot \eta^{e} \cdot\mathcal{L}_\text{reg}, S)$
        \If {$\psi(\nu(W_{e+1}) \odot W_{e+1}) > \psi(\nu(W_\text{best}) \odot W_\text{best})$} 
            \State $W_\text{best} \gets W_{e+1}$ 
        \EndIf
        \State $e \gets e+1$
            
    \EndWhile
\end{algorithmic}
\label{alg:regularTraining}
\end{algorithm}

Regularization losses like the $L_1$-norm ($Lasso$-regression) \cite{Tibshirani1996RegressionSA} or $L_2$-norm \cite{https://doi.org/10.48550/arxiv.1810.12281} are used to prevent overfitting by shrinking the magnitude of weights.
We use this effect in \textit{ART} for sparsification, as weights with low magnitude have low effect on changing the output and thus can be removed with only little impact on $\psi(W)$.

Regularization during training can be expressed as a mixed loss 
\begin{equation}
    \mathcal{L}_\text{total} =  \mathcal{L}_\text{class} + \lambda_\text{init} \cdot \eta^{e} \cdot\mathcal{L}_\text{reg} \ ,
    \label{eq:loss}
\end{equation}
where $\mathcal{L}_\text{class}$ is the classification loss and $\mathcal{L}_\text{reg}$ the regularization loss.
The gradient of $\mathcal{L}_\text{reg}$ shrinks a set of weights to approximately zero and creates a inherent sparse network of an undefined pruning rate~\cite{Tibshirani1996RegressionSA}.
Increasing $\eta$ leverages the regularization $\mathcal{L}_{reg}$ in an ascending manner, but current approaches use a fixed regularization rate $\eta=1$ \cite{NIPS2015_ae0eb3ee, DBLP:journals/corr/MolchanovTKAK16,chen2022sparsity}.

After unregularized training of a dense NN to convergence,
\textit{ART} employs the standard regularization framework and modifies it by setting $\eta>1$ and a low initialisation of $\lambda_\text{init}$.
Subsequently, the regularization loss $\mathcal{L}_\text{reg}$ has almost no effect on $\mathcal{L}_\text{total}$ in the beginning, but starts to shrink weights without much impact on $\mathcal{L}_\text{class}$ to zero. 
However, it allows every weight $w_i$ to potentially get a high magnitude such that $w_i$ is shifted into the sparse NN of highest weights (exploration).
With increasing regularization, the influence of the gradient $\frac{d\mathcal{L}_\text{reg}}{dw_i}$ on $w_i$ increases and is more likely to overcome the gradient $\frac{d\mathcal{L}_\text{class}}{dw_i}$. Regularization impedes proper exploration of small weights by pulling the magnitude to zero. On the other hand, the larger weights need to be exploited to conserve the classification results. Therefore, our increasing regularization continually shifts the exploration/exploitation tradeoff from exploration to exploitation. The method allows reordering weights to find better topologies, but forces to exploit the highest weights regarding the classification task.      
Due to the increasing number of weights that are approximately zero, the dense model converges to a inherently sparse model.
We stop the regularized training if the NN with best pruned weights $\psi(\nu(W_\text{best}) \odot W_\text{best})$ has higher accuracy than with the latest unpruned weights $\psi(W_e)$ and choose $W_\text{best}$ as our candidate for fine-tuning.

The overall training pipeline is defined as follows:
\begin{enumerate}
\item[]\begin{enumerate}
    \item[Step~1:] Pre-train dense model until convergence without regularization.
    \item[Step~2:] Remove weights implicitly using \textit{ART} as described in algorithm \ref{alg:regularTraining}.
    \item[Step~3:] Apply magnitude pruning and fine-tune pruned network until convergence.
\end{enumerate}
\end{enumerate}

\textit{ART} relaxes the iterative \textit{IMP} approach that prunes the least important weights over certain iterations. 
Analogous to the increasing pruning ratio in standard iterative methods, we iteratively increase the amount of weights that are close to zero and thus approximate a binary mask implicitly.

\subsection{HyperSparse Regularization}
\label{sec:SparTan}

The latter Section \ref{sec:regularized_training} describes the process of shrinking weights in $W$ by penalizing with ascending regularization. A drawback of this procedure is that also weights that remain after pruning are penalized by the regularization. This negatively affects the exploitation regarding the main task. Thus, remaining weights should not be penalized. On the other hand, if small weights are strongly penalized, the desired exploration property of dynamic pruning methods to ``grow'' back these elements is restricted. To address this tradeoff between exploitation and exploration, we introduce the sparsity inducing adaptive regularization loss \textit{HyperSparse}.

Incorporating the \textit{Hyper}bolic Tangent function applied on the magnitude denoted as $t(\cdot)=\tanh(|\cdot|)$ for simplicity, the \textit{HyperSparse} loss is defined as  
\begin{equation}
\label{equ:spartan}
\begin{split}
    \mathcal{L}_\text{HS}(W) = & \frac{1}{A} \sum_{i=1}^{|W|} \bigg( |w_i| 
    \sum_{j=1}^{|W|} t(s\cdot w_j)\bigg) - \sum_{i=1}^{|W|} |w_i|
     \\
     & \text{with}\quad  A:= \sum_{w\in W} t(s\cdot w) \\
     & \text{and} \quad \forall w\in W:\quad \frac{dA}{dw} = 0,
\end{split}
\end{equation}
where $A$ is treated as a pseudo-constant in the gradient computation and $s$ is an alignment factor that is described later. The regularization penalizes weights depending on the gradient and can vary for different weights. The gradient of \textit{HyperSparse} with respect to a weight $w_i$ is approximately
\begin{gather}
\label{equ:spartan_grad}
\begin{split}
\frac{d\mathcal{L}_\text{HS}(W)}{dw_i} =\ & \text{sign}(w_i)\cdot 
\frac{t'(s\cdot w_i) \cdot \sum_{j=1}^{|W|}|w_j|}{\sum_{j=1}^{|W|} t(s\cdot w_j)}, \\
  & \quad \text{with}\quad w_i, w_j \in W, \quad t'(\cdot) \in (0, 1].
\end{split}
\end{gather}
The derivative $t'=\frac{dt}{dw_i}$ converges towards $1$ for small magnitudes $|w_i|\approx0$ and towards $0$ for large magnitudes $|w_i|\gg0$. Thus, the second term in Eq.~\eqref{equ:spartan_grad} is adaptive to the weights and highly penalizes small magnitudes, but is breaking down to zero for large ones. Details for the gradient calculation and analysis can be found in the supplementary material, Sec.~D.

The alignment factor $s$ is mandatory to exploit the aforementioned properties for the sparsification task with a specific pruning rate $\kappa$. Since $\mathcal{L}_\text{HS}$ is dependent on the weights magnitude, but there is no determinable value range for weights, our loss $\mathcal{L}_\text{HS}$ is not guaranteed to adapt reasonably to a given $W$. For example, considering a fixed $s=1$ and all weights in $W$ are close to zero, the gradient from Eq.~\eqref{equ:spartan_grad} results into nearly the same value for every weight. Therefore, we adapt $s$ to the smallest weight $|w_\kappa|$ that would remain after magnitude pruning, such that $t'''(s\cdot w_\kappa)=0$, which is the point of inflection of $t'$. According to this alignment, the gradients in Eq.~\eqref{equ:spartan_grad} of remaining weights $|w| \geq |w_\kappa|$ are shifted closer to 1 and are increased for weights $|w| \leq |w_\kappa|$, while adhering a smooth gradient from remaining to removed weights. Moreover, the denominator in Eq.~\eqref{equ:spartan_grad} decreases over time, if more weights in $W$ are close to zero subsequent to ascending regularization. 
The gradient for different weight distributions of a NN based on \textit{HyperSparse} is shown in Fig.~\ref{fig:gradSortetWeightMag} and visualizes the described gradient behavior of adaptive weight regularization. 

\newcommand{\dev}[2]{#1\scriptsize$\pm$#2}
\newcommand{\devB}[2]{\textbf{#1\scriptsize$\pm$#2}}
\newcommand{\devU}[2]{\underline{#1\scriptsize$\pm$#2}}

\begin{table*}
\centering
\small
\setlength{\tabcolsep}{5pt}
\resizebox{\linewidth}{!}{
\begin{tabular}{p{8pt}ccccccc|cccccc} 
 \toprule
 \multirow{2}{*}{$\boldsymbol{\psi}\uparrow$} &  & \multicolumn{6}{c}{\textbf{ResNet-32} $\boldsymbol{\longrightarrow}$} & \multicolumn{6}{c}{\textbf{VGG-19} $\boldsymbol{\longrightarrow}$}\\ 
\cmidrule(rr){3-8}
\cmidrule(rr){9-14}
 & & $\kappa=90\%$ & $\kappa=95\%$ & $\kappa=98\%$ & $\kappa=99\%$ & $\kappa=99.5\%$ & \multicolumn{1}{c}{$\kappa=99.8\%$} & $\kappa=90\%$ & $\kappa=95\%$ & $\kappa=98\%$ & $\kappa=99\%$ & $\kappa=99.5\%$ & $\kappa=99.8\%$ \\
 \midrule

 \multirow{12}{*}{\rotatebox{90}{\textbf{CIFAR-10}}} & No Mask  & \dev{94.70}{0.19} & & & & & & \dev{93.84}{0.12} & & & & &  \\
 
 \cmidrule{3-14}

 & SNIP \cite{lee2018snip} & \dev{92.72}{0.18} & \dev{91.35}{0.15} & \dev{88.02}{0.27} & \dev{83.94}{0.39} & \dev{71.64}{7.46} & \dev{23.72}{21.11} & \dev{93.63}{0.25} & \dev{93.36}{0.20} & \dev{76.12}{21.96} & \dev{10.00}{0.00} & \dev{10.00}{0.00} & \dev{10.00}{0.00}\\

 & GraSP \cite{wang2019picking} & \dev{92.86}{0.19} & \dev{91.80}{0.23} & \dev{89.00}{0.24} & \dev{85.63}{0.28} & \dev{80.25}{0.67} & \dev{62.56}{11.25} & \dev{92.97}{0.04} & \dev{92.79}{0.24} & \dev{92.16}{0.14} & \dev{91.27}{0.15} & \dev{51.45}{38.46} & \dev{10.00}{0.00} \\

 & SRatio \cite{NEURIPS2020_eae27d77}~~& \dev{93.02}{0.17} & \dev{91.85}{0.16} & \dev{88.91}{0.16} & \dev{85.97}{0.22} & \dev{80.73}{0.37} & \dev{64.38}{0.50} & \dev{93.86}{0.19} & \dev{93.58}{0.20} & \dev{92.33}{0.24} & \dev{91.14}{0.21} & \dev{89.14}{0.15} & \dev{43.64}{20.04} \\
 
 & LTH \cite{frankle2018lottery} & \dev{92.68}{0.32} & \dev{91.45}{0.19} & \dev{88.48}{0.15} & \dev{85.99}{0.30} & \dev{81.19}{0.40} & \dev{69.34}{0.43} & \dev{93.71}{0.17} & \dev{93.31}{0.15} & \dev{41.17}{42.76} & \dev{10.00}{0.00} & \dev{10.00}{0.00} & \dev{10.00}{0.00} \\

 & IMP \cite{pmlr-v119-frankle20a} & \devB{94.69}{0.17} & \devB{94.00}{0.18} & \dev{91.35}{0.18} & \dev{87.35}{0.55} & \dev{82.00}{0.34} & \dev{69.12}{0.50} & \devU{93.96}{0.17} & \devB{94.02}{0.07} & \dev{93.48}{0.24} & \dev{91.29}{0.29} & \dev{25.43}{34.50} & \dev{10.00}{0.00} \\
 
 &  RigL \cite{pmlr-v119-evci20a} & \dev{94.21}{0.10} & \dev{93.07}{0.22} & \dev{90.65}{0.17} & \dev{86.50}{0.83} & \dev{62.89}{5.18} & \dev{32.78}{4.11} & \dev{93.48}{0.13} & \dev{92.92}{0.14} & \dev{91.41}{0.15} & \dev{89.08}{0.37} & \dev{84.79}{0.90} & \dev{70.81}{1.10} \\
 
 
 
 \cmidrule{3-14}

 & \textbf{ART} + $\mathcal{L}_1$ & \dev{94.20}{0.16} & \dev{93.14}{0.16} & \dev{91.34}{0.46} & \dev{88.18}{1.07} & \dev{84.52}{1.24} & \dev{79.35}{1.85} & \devB{93.97}{0.13} & \dev{93.82}{0.10} & \devB{93.85}{0.12} & \devU{93.10}{0.23} & \devU{92.17}{0.25} & \dev{90.42}{0.50} \\

 & \textbf{ART} + $\mathcal{L}_2$ & \dev{93.49}{0.21} & \dev{92.91}{0.24} & \dev{89.60}{0.73} & \dev{85.80}{2.38} & \dev{82.24}{0.60} & \dev{71.73}{0.88} & \dev{93.18}{0.18} & \dev{92.65}{0.40} & \dev{79.38}{4.92} & \dev{78.85}{8.74} & \dev{72.68}{2.67} & \dev{56.28}{26.33} \\

 &  \textbf{ART + $\boldsymbol{\mathcal{L}_\text{HS}}$ (no preTrain)} & \dev{93.13}{0.13} & \dev{92.85}{0.18} & \devU{91.79}{0.14} & \devU{90.79}{0.30} & \devU{89.01}{0.21} & \devB{84.64}{0.51} & \dev{93.58}{0.12} & \dev{93.53}{0.09} & \dev{93.15}{0.12} & \dev{92.56}{0.08} & \dev{92.12}{0.13} & \devU{91.24}{0.09} \\

 & \textbf{ART + $\boldsymbol{\mathcal{L}_\text{HS}}$} & \devU{94.22}{0.20} & \devU{93.76}{0.18} & \devB{92.69}{0.22} & \devB{91.16}{0.28} & \devB{89.35}{0.23} & \devU{84.45}{0.55} & \dev{93.93}{0.20} & \devU{93.83}{0.10} & \devU{93.75}{0.23} & \devB{93.51}{0.15} & \devB{92.91}{0.10} & \devB{91.62}{0.19} \\
 
 \midrule
 
 \multirow{12}{*}{\rotatebox{90}{\textbf{CIFAR-100}}} & No Mask  & \dev{74.60}{0.14} & & & & & & \dev{72.88}{0.34} & & & & & \\
 
 \cmidrule{3-14}

 & SNIP \cite{lee2018snip} & \dev{69.78}{0.22} & \dev{65.54}{0.26} & \dev{53.20}{0.30} & \dev{37.45}{1.42} & \dev{14.76}{3.35} & \dev{04.52}{2.16} & \dev{72.76}{0.20} & \dev{71.50}{0.27} & \dev{25.34}{9.16} & \dev{1.00}{0.00} & \dev{1.00}{0.00} & \dev{1.00}{0.00} \\

 & GraSP \cite{wang2019picking} & \dev{69.64}{0.38} & \dev{66.84}{0.14} & \dev{59.59}{0.30} & \dev{49.42}{1.04} & \dev{36.46}{2.73} & \dev{15.62}{3.20} & \dev{71.10}{0.13} & \dev{70.39}{0.17} & \dev{68.25}{0.45} & \dev{65.84}{0.36} & \dev{59.56}{0.47} & \dev{1.10}{0.10} \\

 & SRatio \cite{NEURIPS2020_eae27d77} & \dev{69.80}{0.18} & \dev{67.08}{0.41} & \dev{60.44}{0.32} & \dev{51.60}{0.63} & \dev{38.57}{0.75} & \dev{18.35}{0.97} & \dev{72.84}{0.32} & \dev{71.67}{0.19} & \dev{68.84}{0.38} & \dev{65.00}{0.22} & \dev{51.16}{2.67} & \dev{1.02}{0.04} \\
 
 & LTH \cite{frankle2018lottery} & \dev{69.23}{0.31} & \dev{66.80}{0.49} & \dev{60.28}{0.10} & \dev{51.92}{0.11} & \dev{40.18}{0.28} & \dev{20.31}{1.63} & \dev{72.55}{0.27} & \dev{70.46}{0.26} & \dev{9.80}{15.98} & \dev{1.00}{0.00} & \dev{1.00}{0.00} & \dev{1.00}{0.00} \\

 & IMP \cite{pmlr-v119-frankle20a} & \devU{73.91}{0.37} & \dev{71.21}{0.36} & \dev{64.67}{0.29} & \dev{55.89}{0.34} & \dev{41.53}{0.74} & \dev{14.97}{0.69} & \devB{73.92}{0.33} & \devB{73.77}{0.32} & \dev{70.99}{0.34} & \dev{4.03}{4.69} & \dev{1.00}{0.00} & \dev{1.00}{0.00} \\

 &  RigL \cite{pmlr-v119-evci20a} & \dev{73.09}{0.29} & \dev{71.46}{0.37} & \dev{64.46}{0.36} & \dev{45.58}{1.78} & \dev{21.80}{1.54} & \dev{8.47}{4.24} & \dev{72.00}{0.24} & \dev{70.42}{0.30} & \dev{67.48}{0.36} & \dev{63.31}{0.51} & \dev{55.56}{1.33} & \dev{24.57}{13.21} \\

 \cmidrule{3-14}

 & \textbf{ART} + $\mathcal{L}_1$ & \dev{73.16}{0.45} & \dev{70.98}{0.48} & \dev{66.10}{0.76} & \dev{59.36}{2.08} & \dev{50.50}{3.43} & \dev{37.43}{1.64} & \dev{73.16}{0.20} & \devU{72.80}{0.20} & \devU{71.23}{0.25} & \devU{69.18}{0.22} & \devU{65.71}{0.63} & \dev{59.08}{1.07} \\

 & \textbf{ART} + $\mathcal{L}_2$ & \dev{71.39}{0.60} & \dev{68.21}{1.25} & \dev{58.49}{3.93} & \dev{56.61}{0.88} & \dev{47.11}{1.00} & \dev{28.73}{1.18} & \dev{61.54}{3.91} & \dev{55.22}{7.34} & \dev{44.42}{6.14} & \dev{39.40}{4.78} & \dev{26.94}{23.75} & \dev{29.78}{16.25} \\
 
 &  \textbf{ART + $\boldsymbol{\mathcal{L}_\text{HS}}$ (no preTrain)} & \dev{72.49}{0.35} & \devU{71.57}{0.36} & \devU{69.08}{0.12} & \devU{65.48}{0.28} & \devU{59.49}{0.53} & \devB{48.63}{0.66} & \dev{71.49}{0.42} & \dev{70.24}{0.67} & \dev{68.57}{0.38} & \dev{67.59}{0.47} & \dev{65.59}{0.17} & \devU{61.66}{0.50}\\

 & \textbf{ART + $\boldsymbol{\mathcal{L}_\text{HS}}$} & \devB{74.08}{0.13} & \devB{72.85}{0.31} & \devB{70.08}{0.37} & \devB{65.86}{0.26} & \devB{59.58}{0.26} & \devU{48.31}{0.53} & \devU{73.23}{0.24} & \dev{72.70}{0.41} & \devB{71.97}{0.13} & \devB{70.83}{0.23} & \devB{69.02}{0.36} & \devB{64.53}{0.24}\\
 
 \midrule

 \multirow{12}{*}{\rotatebox{90}{\textbf{TinyImageNet}}} & No Mask  & \dev{62.87}{0.27} & & & & & & \dev{61.41}{0.12} & & & & & \\\cmidrule{3-14}

 & SNIP \cite{lee2018snip} & \dev{55.23}{0.47} & \dev{48.78}{0.40} & \dev{34.93}{0.83} & \dev{23.20}{1.41} & \dev{12.25}{1.50} & \dev{3.19}{1.52} & \dev{61.47}{0.16} & \dev{59.00}{0.20} & \dev{4.77}{4.23} & \dev{0.50}{0.00} & \dev{0.50}{0.00} & \dev{0.50}{0.00} \\

 & GraSP \cite{wang2019picking} & \dev{56.16}{0.25} & \dev{51.52}{0.47} & \dev{40.32}{2.24} & \dev{28.41}{1.26} & \dev{15.81}{2.30} & \dev{4.29}{3.73} & \dev{60.50}{0.08} & \dev{58.97}{0.14} & \dev{56.70}{0.12} & \dev{53.12}{0.49} & \dev{43.76}{0.40} & \dev{0.51}{0.03} \\

 & SRatio \cite{NEURIPS2020_eae27d77} & \dev{55.19}{0.35} & \dev{51.70}{0.48} & \dev{44.04}{0.36} & \dev{34.14}{0.12} & \dev{8.31}{1.26} & \dev{1.98}{0.29} & \dev{61.21}{0.19} & \dev{59.10}{0.32} & \dev{55.94}{0.24} & \dev{51.13}{0.34} & \dev{39.76}{0.32} & \dev{0.50}{0.00} \\
 
 &  LTH \cite{frankle2018lottery} & \dev{55.72}{0.22} & \dev{52.22}{0.48} & \dev{43.73}{0.85} & \dev{33.22}{0.39} & \dev{20.78}{0.40} & \dev{7.65}{0.58} & \dev{59.91}{0.59} & \dev{58.74}{0.43} & \dev{56.38}{0.16} & \dev{54.02}{0.60} & \dev{46.78}{0.81} & \dev{2.89}{2.32} \\
 
 &  IMP \cite{pmlr-v119-frankle20a} & \devU{60.71}{0.24} & \dev{56.97}{0.26} & \dev{47.29}{0.57} & \dev{33.21}{0.11} & \dev{8.59}{0.67} & \dev{2.40}{0.34} & \devB{62.42}{0.32} & \dev{61.28}{0.23} & \dev{57.39}{0.10} & \dev{54.26}{0.26} & \dev{47.19}{0.28} & \dev{3.10}{0.96} \\
 
 &  RigL \cite{pmlr-v119-evci20a} & \dev{59.29}{0.21} & \dev{55.53}{0.16} & \dev{44.72}{1.34} & \dev{26.07}{1.59} & \dev{8.76}{0.30} & \dev{4.51}{0.37} & \dev{61.47}{0.29} & \devB{61.69}{0.41} & \devU{59.41}{0.53} & \dev{54.59}{0.68} & \dev{47.11}{0.62} & \dev{20.81}{0.99} \\
 
 \cmidrule{3-14}

 & \textbf{ART} + $\mathcal{L}_1$  & \dev{58.00}{0.49} & \dev{55.30}{0.94} & \dev{46.59}{0.45} & \dev{39.34}{1.32} & \dev{29.80}{3.53} & \dev{18.06}{2.89} & \dev{61.29}{0.24} & \dev{60.21}{0.31} & \dev{57.04}{0.53} & \devU{54.61}{0.77} & \dev{51.27}{1.80} & \dev{43.59}{1.33} \\
 
  & \textbf{ART} + $\mathcal{L}_2$ & \dev{56.94}{0.76} & \dev{51.40}{0.99} & \dev{43.03}{2.23} & \dev{35.19}{1.42} & \dev{24.62}{1.96} & \dev{7.79}{0.59} & \dev{60.62}{0.69} & \dev{51.10}{5.94} & \dev{47.96}{7.73} & \dev{45.90}{8.76} & \dev{30.32}{17.94} & \dev{5.55}{11.30} \\
 
 &  \textbf{ART + $\boldsymbol{\mathcal{L}_\text{HS}}$ (no preTrain)} & \dev{57.96}{0.39} & \devU{57.01}{0.34} & \devU{53.27}{0.32} & \devU{47.32}{0.46} & \devU{40.53}{0.26} & \devB{28.96}{0.69} & \dev{60.95}{0.27} & \dev{59.67}{0.17} & \dev{56.72}{0.48} & \dev{53.79}{0.33} & \devU{51.66}{0.19} & \devU{47.49}{0.17}\\

 & \textbf{ART + $\boldsymbol{\mathcal{L}_\text{HS}}$} & \devB{60.97}{0.18} & \devB{58.78}{0.28} & \devB{53.92}{0.14} & \devB{47.97}{0.42} & \devB{40.68}{0.82} & \devU{28.95}{0.52} & \devU{61.55}{0.24} & \devU{61.36}{0.31} & \devB{59.79}{0.25} & \devB{58.01}{0.21} & \devB{55.34}{0.22} & \devB{49.44}{0.18}\\

\bottomrule
\end{tabular}}
\vspace{-0.2cm}
\caption{Classification accuracy of sparse NNs for varying pruning rates $\kappa$ based on our proposed method \textit{ART} with $\mathcal{L}_1$, $\mathcal{L}_2$, and \textit{HyperSparse} regularization  $\mathcal{L}_\text{HS}$ compared to dense models, and masks obtained by \textit{SNIP}~\cite{lee2018snip}, \textit{GraSP}~\cite{wang2019picking}, \textit{SRatio}~\cite{NEURIPS2020_eae27d77}, \textit{LTH}~\cite{frankle2018lottery}, \textit{IMP}~\cite{pmlr-v119-frankle20a} and \textit{RigL}~\cite{pmlr-v119-evci20a}. The best accuracy per configuration is highlighted, the second is underlined. It shows that our method outperforms \textit{IMP} significantly in the domain of high sparsity. We recommend the pdf version and zooming in.}
\vspace{-0.6cm}
\label{table:prune_methods}
\end{table*}

\newcommand{\devE}[2]{#1\scriptsize$\pm$#2}
\newcommand{\devBE}[2]{\textbf{#1\scriptsize$\pm$#2}}

\begin{table}
\vspace{+0.15cm}
\centering
\setlength{\tabcolsep}{4pt}
\small
\resizebox{\linewidth}{!}{
\begin{tabular}{p{35pt} p{20pt} cccc|ccc} 
 \toprule
 \multirow{2}{*}{\textbf{\#Epochs} $\downarrow$} & & & \multicolumn{3}{c}{\textbf{ResNet-32} $\boldsymbol{\longrightarrow}$} & \multicolumn{3}{c}{\textbf{VGG-19 $\boldsymbol{\longrightarrow}$}}\\ 
 \cmidrule(rr){3-6}\cmidrule(rr){7-9}

  &  & $\kappa$: & 90\% & 98\% & 99.5\% & 90\% & 98\% & 99.5\% \\
 \midrule
\multirow{3}{*}{\hfil \rotatebox{90}{\textbf{\shortstack{CIFAR-~\\10}}} \hfil}
 & $\mathcal{L}_1$    & & \devE{34.2}{3.1} & \devE{68.2}{4.8} & \devE{94.2}{8.0}  & \devE{5.8}{0.4}   & \devE{24.2}{2.4}  & \devE{56.0}{3.0} \\
 & $\mathcal{L}_2$    & & \devE{75.6}{63.6} & \devE{49.6}{92.45} & \devE{116.6}{19.5} & \devE{116.2}{5.3} & \devE{175.8}{15.8} & \devE{178.4}{9.7} \\
 & \textbf{$\boldsymbol{\mathcal{L}_\text{HS}}$}  & & \devBE{26.6}{1.8} & \devBE{55.4}{2.3} & \devBE{77.8}{1.9}  & \devBE{4.0}{0.0}   & \devBE{18.2}{1.6}  & \devBE{42.8}{1.6} \\
 
 \midrule
 \multirow{3}{*}{\hfil \rotatebox{90}{\textbf{\shortstack{CIFAR-~\\100}}} \hfil}
 & $\mathcal{L}_1$       & & \devE{53.2}{2.8} & \devE{77.5}{4.1} & \devE{101.3}{9.6} & \devE{11.2}{0.4} & \devE{47.7}{2.1} & \devE{66.5}{4.6} \\
 & $\mathcal{L}_2$       & & \devE{112.0}{7.6} & \devE{141.8}{17.0} & \devE{120.0}{11.4} & \devE{153.2}{2.9} & \devE{168.8}{6.9} & \devE{75.4}{122.7} \\
 & \textbf{$\boldsymbol{\mathcal{L}_\text{HS}}$}  & & \devBE{39.2}{0.84} & \devBE{63.4}{0.5} & \devBE{88.2}{0.8}  & \devBE{8.2}{0.4}   & \devBE{35.8}{0.4}  & \devBE{55.8}{0.4} \\
 \midrule 
 \multirow{3}{*}{\hfil  \rotatebox{90}{\textbf{\shortstack{Tiny-\\Image-\\Net}}} \hfil} 
 & $\mathcal{L}_1$       & & \devE{52.0}{3.4} & \devE{81.6}{3.5} & \devE{101.2}{10.8} & \devE{20.8}{0.4} & \devE{43.2}{3.5} & \devE{59.0}{8.0} \\
 & $\mathcal{L}_2$       & & \devE{110.2}{7.3} & \devE{129.8}{11.6} & \devE{93.2}{45.5} & \devE{27.6}{81.1} & \devE{148.4}{22.3} & \devE{107.4}{94.6}
 \\
 & \textbf{$\boldsymbol{\mathcal{L}_\text{HS}}$}  & & \devBE{36.6}{1.3} & \devBE{67.8}{3.8} & \devBE{100.0}{9.3}  & \devBE{14.4}{0.5}   & \devBE{34.0}{0.0}  & \devBE{52.3}{1.5} \\
 \bottomrule
\end{tabular}}
\vspace{-0.2cm}
\caption{Number of epochs with regularization to obtain the final mask, evaluated for multiple datasets, network topologies, and pruning rates $\kappa$. It shows that our \textit{HyperSparse} $\mathcal{L}_\text{HS}$ loss reduces the training time significantly. }
\vspace{-0.6cm}
\label{table:trainEpochs}
\end{table}

\section{Experiments}

This section presents experiments showing that our proposed method \textit{ART} outperforms comparable methods, especially in extreme high sparsity regimes. 
Our experimental setup is described in Sec.~\ref{sec:experimentalSetup}.
In the subsequent section, we show that \textit{HyperSparse} has a large positive impact on the optimization time and classification accuracy. This improvement is explained by analyzes of the tradeoff between exploration and exploitation, the gradient and weight distribution in Sec.~\ref{sec:explorationExploitation} and \ref{sec:reorderingWeights}. Finally, we analyze and discuss the compression behaviour during regularized training and derive further insights about highest magnitude weights in Sec.~\ref{sec:whatDoNetworksCompressFirst}.

\subsection{Experimental Setup}
\label{sec:experimentalSetup}

We evaluate \textit{ART} on the datasets CIFAR-10/100~\cite{krizhevsky2009learning} and TinyImageNet~\cite{deng2009imagenet} to cover different complexities, given by a varying number of class labels.
Furthermore, we use different model complexities, where ResNet-32~\cite{He_2016_CVPR} is a simple model with 1.8~M parameters and VGG-19~\cite{simonyan2014very} is a complex model with 20~M parameters.
Note that we use the implementation given in \cite{NEURIPS2020_eae27d77}.
As explained in Sec.~\ref{sec:regularized_training}, we group our training in 3 steps.
First we train our model for 60 epochs until convergence (step~1), using a constant learning rate of $0.1$.
In the following regularization step, we initialize the regularization with $\lambda_\text{init}=5\cdot 10^{-6}$, $\eta=1.05$, and use the same learning rate as used in pre-training.
The fine-tuning-step (step~3) is similar to~\cite{NEURIPS2020_eae27d77}, as we train for 160 epochs in CIFAR-10/100 and for 300 epochs on TinyImageNet, using a learning rate of $0.1$ and apply a multiplied decay of 0.1 at 2/4 and 3/4 of the total number of epochs.
We also adapt the batch size of 64 and weight-decay of $10^{-4}$.
All experiments are averaged over 5 runs.

We compare our method \textit{ART} to \textit{SNIP}\cite{lee2018snip}, \textit{Grasp}~\cite{wang2019picking}, \textit{SRatio}~\cite{NEURIPS2020_eae27d77}, and \textit{LTH} \cite{frankle2018lottery} similar as done in \cite{NEURIPS2020_eae27d77,Verma_2023_WACV}.
In addition we evaluate \textit{IMP}~\cite{pmlr-v119-frankle20a} and \textit{RigL}~\cite{pmlr-v119-evci20a} as dynamic pruning methods.
For comparability, all competitors in our experiments are trained with the same setup as given in the fine-tuning-step.
To improve the performance of \textit{RigL}, we extend the training duration by 360 epochs.
Further details are given in the supplementary material, Sec.~A.

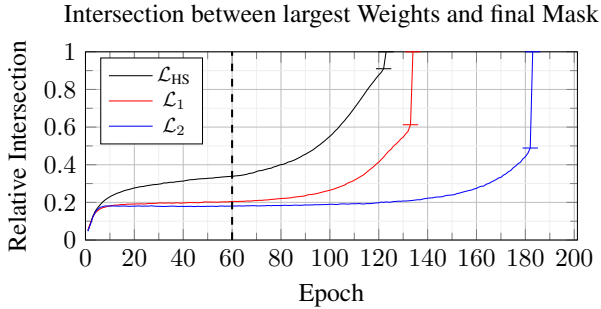
\begin{figure}[t]
\centering
\resizebox{\columnwidth}{!}{
\newcommand{\fileMaskIntersec}{ablation_results/diffMask/relMaskEq_dataset=cifar100_model=resnet32_pruneRate=0.98}

\begin{tikzpicture}
\begin{axis}[
    title=Intersection between largest Weights and final Mask,
    scaled ticks=false, 
    log ticks with fixed point,
    tick label style={/pgf/number format/fixed},
    xmin = 0, 
    ymin = 0.0, ymax = 1.0,
    xtick distance = 20,
    ytick distance = 0.2,
    grid = both,
    minor tick num = 1,
    major grid style = {lightgray},
    minor grid style = {lightgray!25},
    width = 1\columnwidth,
    height = 0.5\linewidth,
    xlabel = {Epoch},
    ylabel = {Relative Intersection},
    legend pos = north west,
    legend style = {
      anchor=north west,
      legend columns=1,
      nodes={scale=0.8, transform shape}},
]

\addplot[black] table[x=epoch,y=mDiff,col sep=comma] {\fileMaskIntersec_HyperSparse.txt};

\addplot[red] table[x=epoch,y=mDiff,col sep=comma] {\fileMaskIntersec_L1.txt};

\addplot[blue] table[x=epoch,y=mDiff,col sep=comma] {\fileMaskIntersec_L2.txt};

\addplot[skip coords between index={0}{121}, only marks, mark options={draw=black, mark=-, mark size=3.0pt}]
	table[x=epoch,y=mDiff,col sep=comma] {\fileMaskIntersec_HyperSparse.txt};
	
\addplot[skip coords between index={0}{132}, only marks, mark options={draw=red, mark=-, mark size=3.0pt}]
	table[x=epoch,y=mDiff,col sep=comma] {\fileMaskIntersec_L1.txt};

\addplot[skip coords between index={0}{181}, only marks, mark options={draw=blue, mark=-, mark size=3.0pt}]
	table[x=epoch,y=mDiff,col sep=comma] {\fileMaskIntersec_L2.txt};
    
\draw [dashed, thick] (60,0) -- (60,1);

\legend{$\mathcal{L}_\text{HS}$, $\mathcal{L}_1$, $\mathcal{L}_2$}
\end{axis}
\end{tikzpicture}}
\vspace{-20pt}
\caption{Intersection of the set of weights with highest magnitude during training and the final mask measured during \textit{ART} with ResNet-32, CIFAR-100, pruning rate $\kappa=98\%$ and different regularization losses. Horizontal bars mark the intersection one epoch before pruning and the dashed line at epoch~$60$ indicates the start of regularization. Our \textit{HyperSparse} loss reduces the optimization time and the high intersection before pruning suggests a higher stability during regularization, which leads to better exploitation.}
\vspace{-5pt}
\label{fig:mask_intersection}
\end{figure}

\subsection{Sparsity Level}
\label{sec:sparsityLevel}

In this section, we compare the performances of \textit{ART} to other methods on different sparsity levels $\kappa \in \{ 90\%, 95\%, 98\%, 99\%, 99.5\%, 99.8\%\}$, using different datasets and models.
To demonstrate the advantages of our novel regularization loss, we additionally substitute \textit{HyperSparse} with $\mathcal{L}_1$~\cite{Tibshirani1996RegressionSA} and $\mathcal{L}_2$~\cite{https://doi.org/10.48550/arxiv.1810.12281}.
Table~\ref{table:prune_methods} shows the resulting accuracies with standard deviations.

Our method \textit{ART} combined with \textit{HyperSparse} outperformes the methods \textit{SNIP}~\cite{lee2018snip}, \textit{Grasp}~\cite{wang2019picking}, \textit{SRatio}~\cite{NEURIPS2020_eae27d77}, \textit{LTH} \cite{frankle2018lottery} and \textit{RigL}~\cite{pmlr-v119-evci20a} on all sparsity levels.
Considering the high sparsity of $99\%$, $99.5\%$ and $99.8\%$, all competitors drop drastically in accuracy, even to the minimal classification bound of random prediction for SNIP and \textit{LTH} using VGG-19.
However, \textit{ART} is able keep high accuracy even on extreme high sparsity levels.
In comparison to the regularization losses $\mathcal{L}_1$ and $\mathcal{L}_2$, our \textit{HyperSparse} loss achieves higher accuracy in nearly all settings and even minimizes the variance.
If we skip the Pre-train-step (step~1) of \textit{ART}, the performance slightly drops.
However, \textit{ART} without pre-training still has good results.

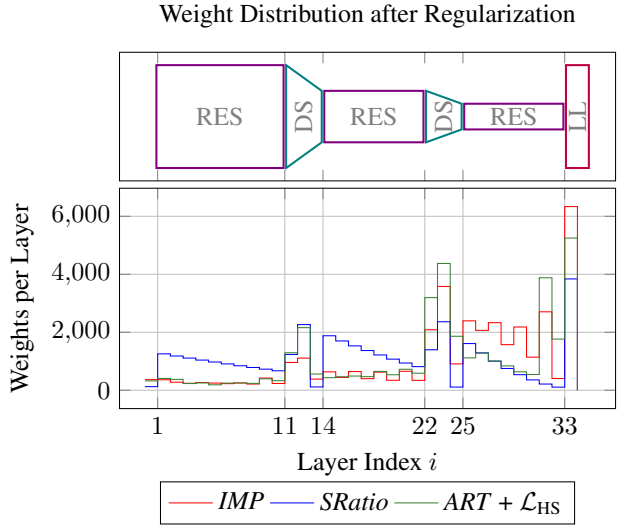
\begin{figure}[t]
\centering
\resizebox{\linewidth}{!}{
\begin{tikzpicture}

\begin{groupplot}[group style={group size= 1 by 2},
                    height = 0.2\linewidth,width=\columnwidth]
    \pgfplotsset{/pgfplots/group/.cd, vertical sep=0.1cm}
    \nextgroupplot[
        title=Weight Distribution after Regularization,
        width = 1\columnwidth,
        height = 0.4\linewidth,
        xmin = -2, xmax = 37,
        ymin = 0., ymax = 1.0,
        xticklabels=\empty,
        yticklabels=\empty,
        xtick = {1, 11, 14 ,22 ,25, 33},
        ytick={-10, 10}
    ]
    \draw[violet, thick] (0.9,0.1) rectangle (10.9,0.9) node[gray, pos=.5] {RES};
    
    \draw[teal, thick] (11.1,0.1) -- (13.9,0.3) -- (13.9,0.7) -- (11.1,0.9) -- cycle node[gray, anchor=west] at (11,.5) {\rotatebox{90}{DS}}; 
    
    \draw[violet, thick] (14.1,0.3) rectangle (21.9,0.7) node[gray, pos=.5] {RES};
    
    \draw[teal, thick] (22.1,0.3) -- (24.9,0.4) -- (24.9,0.6) -- (22.1,0.7) -- cycle node[gray, anchor=west] at (22,.5) {\rotatebox{90}{DS}}; 
    
    \draw[violet, thick] (25.1,0.4) rectangle (32.9,0.6) node[gray, pos=.5] {RES};
    
    \draw[purple, thick] (33.1,0.1) rectangle (34.9,0.9) node[gray, pos=.5] {\rotatebox{90}{LL}};  
    
    \nextgroupplot[
        bar width=3pt,
        scaled ticks=false, 
        log ticks with fixed point,
        tick label style={/pgf/number format/fixed},
        xtick = {1, 11, 14 ,22 ,25, 33},
        xmin = -2, xmax = 37,
        grid = both,
        minor tick num = 0,
        major grid style = {lightgray},
        minor grid style = {lightgray!50},
        width = 1\columnwidth,
        height = 0.55\linewidth,
        xlabel = {Layer Index $i$},
        ylabel = {Weights per Layer},
        legend style = {
          at={(0.5,-0.33)},
          anchor=north,
          legend columns=5},
    ]
    
    \addplot[red, const plot] table [x=l_numb,y=p_keep_mean, col sep=comma] {ablation_results/param_per_layer/PpL_method=IMP_dataset=cifar100_model=resnet32_pruneRate=0.98.txt};
    \addplot[blue, const plot] table [x=l_numb,y=p_keep_mean, col sep=comma] {ablation_results/param_per_layer/PpL_method=smartRatio_dataset=cifar100_model=resnet32_pruneRate=0.98.txt};
    \addplot[OliveGreen, const plot] table [x=l_numb,y=p_keep_mean, col sep=comma] {ablation_results/param_per_layer/PpL_method=RegularMask_dataset=cifar100_model=resnet32_pruneRate=0.98.txt};

    \legend{\textit{IMP}, \textit{SRatio}, \textit{ART} + $\mathcal{L}_\text{HS}$}

\end{groupplot}
\end{tikzpicture}}
\vspace{-15pt}
\caption{Distribution of weights per layer after pruning in a ResNet-32 model that is trained on CIFAR-100 with pruning rate $\kappa=98\%$. 
Layer index $i$ describes the execution order.
We group the model in residual blocks (RES), downsampling blocks (DS) and the linear layer (LL). Our method distributes the weights comparable to 
\textit{IMP}~\cite{pmlr-v119-frankle20a}, but it has more weights in the downsampling layers. 
}
\label{fig:weight_distribution}
\vspace{-10pt}
\end{figure}

Moreover, we present the number of trained epochs for the regularization phase (step~2) in Tab.~\ref{table:trainEpochs}.
In almost all cases, \textit{HyperSparse} requires less epochs to terminate compared to $\mathcal{L}_1$ and $\mathcal{L}_2$ and converges faster to a well performing sparse model. 
As a second aspect, \textit{ART} dynamically varies the training-length to the sparsity level, model and data complexity.
Thus, \textit{ART} trains longer if higher sparsity is required or the model has more parameters and is more complex like VGG-19.
In comparison of the two datasets CIFAR-10 and CIFAR-100, which have the same number of training samples and thus the same number of optimization steps per epoch, \textit{ART} extends the training-length for the more complex classification problem in CIFAR-100.

\textit{ART} trains the model for 60 epochs in pre-training (step~1) and 160 epochs in fine-tuning (step~3).
Considering the dynamic training-length in step~2, the epochs of \textit{ART} using $\mathcal{L}_\text{HS}$ sum up from $226.2$ to $301.2$ epochs in mean.
In comparison, iterative pruning methods are computationally much more expensive, since each model is trained multiple times.
For example, IMP~\cite{pmlr-v119-frankle20a} requires 860 epochs on CIFAR-10/100 in our experiments.


\subsection{Exploration and Exploitation aware Gradient}
\label{sec:explorationExploitation}
\begin{figure*}[ht]
    \centering
     \begin{subfigure}[c]{0.28\textwidth}
         \centering
         \includegraphics[trim={0.3cm 0 0.2cm 0.6cm},clip, width=\textwidth, height=0.85\textwidth]{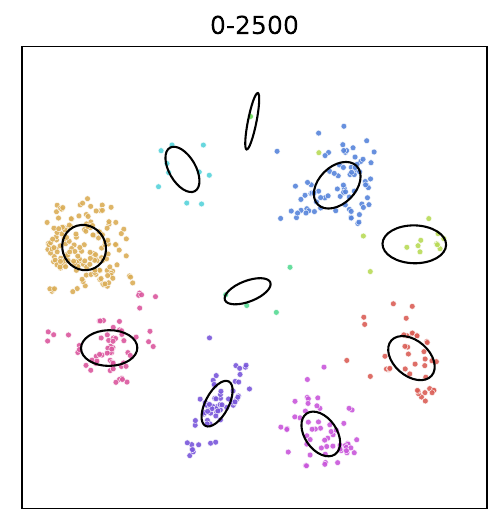}
     \vspace{-20pt}
     \caption{Dense Model ($\kappa=0\%$)}
     \end{subfigure}
     \begin{subfigure}[c]{0.28\textwidth}
         \centering
         \includegraphics[trim={0.3cm 0 0.2cm 0.6cm},clip,width=\textwidth, height=0.85\textwidth]{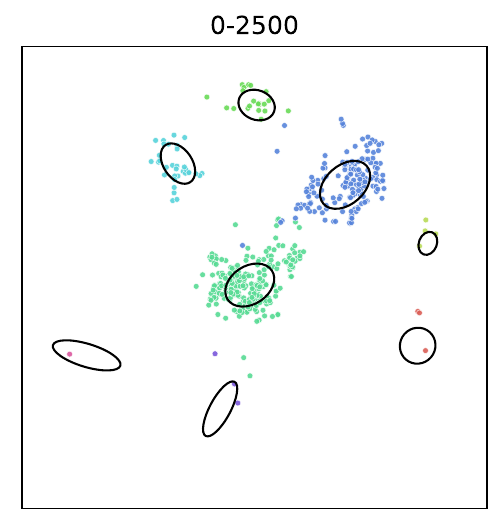}
     \vspace{-20pt}
     \caption{Low Sparsity ($\kappa=90\%$)}
     \end{subfigure}
     \begin{subfigure}[c]{0.28\textwidth}
         \centering
         \includegraphics[trim={0.3cm 0 0.2cm 0.6cm},clip,width=\textwidth, height=0.85\textwidth]{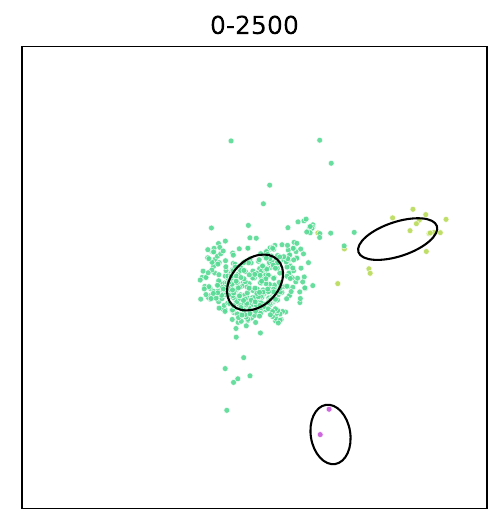}
     \vspace{-20pt}
     \caption{High Sparsity ($\kappa=99.8\%$)}
     \end{subfigure}
     \hfill
     \begin{subfigure}[c]{0.13\textwidth}
         \centering         
         \includegraphics[width=\textwidth]{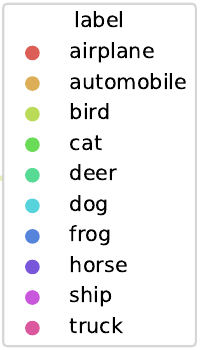}
         \vspace{1pt}
     \end{subfigure}
    \vspace{-5pt}    \caption{First 5\% CIFAR-10 samples that are compressed into the remaining highest weights after pruning with $\kappa \in \{0\%, 90\%, 99.8\%\}$ deduced by the CP-metric. While dense networks learn samples approximately uniform-distributed over classes, the highest weights compress decision rules only for a subset of classes in the early learning stage. Note that we sampled by factor 10 for visualization purposes and ellipses represent the double standard deviation of cluster centers.}
    \label{fig:learning_clusters}
    \vspace{-15pt}
\end{figure*}

The training-schedule of \textit{ART} allows to explore new topologies of sparse networks, while compressing the dense network into the remaining weights that are exploited to minimize the loss $\mathcal{L}_\text{class}$.
To reduce the tradeoff between exploration and exploitation, our regularization loss \textit{HyperSparse} penalizes small weights with a higher regularization and forces the most weights to be close to zero, while preserving the magnitude of weights that remain after pruning. To highlight the beneficial behaviour of \textit{HyperSparse}, this section visualizes and analyzes the gradient. Fig.~\ref{fig:gradSortetWeightMag} shows the values and the corresponding gradients of all weights, sorted by the weights magnitude.
Note that we only focus on the second step of \textit{ART}, where the regularization is incorporated.
Epoch $0$ represents the first epoch using regularization.
In the lower subfigure, we observe that the gradient of \textit{HyperSparse} with respect to weights larger than $|w_\kappa|$ is closer to 0 than for smaller weights. In comparison, $\mathcal{L}_1$ remains constantly 1 for all weights.
The effect of increasing regularization of small weights is stronger for networks with more weights close to zero and therefore amplifies over time, since increasing regularization shrinks the weights magnitude. For example, epoch 40 shows higher gradients for small weights compared to epoch 0, while having more weights with lower magnitude.
The pruning-rate $\kappa$ dependent $\mathcal{L}_\text{HS}$ increases the gradient for small weights $|w|<|w_\kappa|$ over time but conserves the low gradient of larger weights $|w|>|w_\kappa|$ approximately at 0 to favor exploitation.
During optimization, the gradient remains smooth and increases slowly for weights that are smaller, but close to $|w_\kappa|$. This favors exploration in the domain of weights close to $w_\kappa$. 
Therefore, the model becomes inherently sparse and the behaviour shifts continuously from exploration to exploitation.

\begin{table}
\centering
\small
\resizebox{\linewidth}{!}{
\begin{tabular}{cc|cccccc} 
 \toprule
 \multirow{2}{*}{\textbf{CP}} &\multicolumn{1}{c}{\multirow{2}{*}{\textbf{\shortstack[c]{Human\\Label\\ Errors}}}} & \multicolumn{6}{c}{\textbf{CIFAR-10 Class}} \\ \cmidrule{3-8}

 &  \multicolumn{1}{c}{}&
 deer &
 bird &
 cat &
 truck &
 airplane &
 horse\\
 \midrule
 
  \multirow{4}{*}{\rotatebox{90}{\shortstack[c]{\textbf{Dense}\\\textbf{Model}\\\textbf{(0\%)}}}} 
& 0 & 0.482 & 0.554 & 0.639 & 0.400 & 0.452 & 0.429 \\
& 1 & 0.548 & 0.632 & 0.722 & 0.479 & 0.547 & 0.493 \\
& 2 & 0.653 & 0.741 & 0.808 & 0.588 & 0.676 & 0.710 \\
& 3 & 0.760 & 0.823 & 0.862 & 0.695 & 0.783 & 0.769  \\
\midrule
 
 \multirow{4}{*}{\rotatebox{90}{\shortstack[c]{\textbf{Low}\\\textbf{Sparsity}\\\textbf{(90\%)}}}}
& 0 & 0.166 & 0.499 & 0.494 & 0.567 & 0.580 & 0.524  \\
& 1 & 0.217 & 0.573 & 0.601 & 0.613 & 0.671 & 0.570 \\
& 2 & 0.292 & 0.678 & 0.721 & 0.676 & 0.790 & 0.737 \\
& 3 & 0.392 & 0.772 & 0.796 & 0.710 & 0.841 & 0.808 \\
\midrule
 
   \multirow{4}{*}{\rotatebox{90}{\shortstack[c]{\textbf{High}\\\textbf{Sparsity}\\\textbf{(99.8\%)}}}} 
& 0 & 0.056 & 0.296 & 0.387 & 0.622 & 0.798 & 0.824  \\
& 1 & 0.061 & 0.317 & 0.411 & 0.647 & 0.835 & 0.835  \\
& 2 & 0.069 & 0.342 & 0.439 & 0.677 & 0.882 & 0.893  \\
& 3 & 0.081 & 0.363 & 0.469 & 0.699 & 0.902 & 0.929  \\

\bottomrule
\end{tabular}}
\vspace{-3pt}
\caption{\textit{Compression Position} (see Sec.~\ref{sec:whatDoNetworksCompressFirst}) for dense NNs (during pre-training) and $\kappa$ pruned NNs (during regularization) for six CIFAR-10 classes. Samples of a class are split into 4 subsets according to the number of human label errors in CIFAR-N to indicate the difficulty. In sparse networks, different classes are compressed at different times and difficult samples are compressed later. All classes and pruning rates can be found in the supplementary material, Tab.~2.}
\label{table:human_score_position}
\vspace{-10pt}
\end{table}

\subsection{Reordering Weights}
\label{sec:reorderingWeights}
We use the regularization loss with ascending leverage to find a reasonable set of weights, that remain after pruning.
We implicitly do this by shrinking small weights close to zero.
During training, weights are reordered and thus can change the membership from the set of pruned to remaining weights, and vice versa.
We analyze the reordering procedure in  Fig.~\ref{fig:mask_intersection}, which shows the intersection of the intermediate and final mask over all epochs, using different regularization losses in \textit{ART}. 
The model is pre-trained to convergence without regularization for the first 60 epochs (step~1) and with regularization in further epochs (step 2). Fine-tuning is not visualized (step~3).
After pre-training, the highest weights only intersects up to $20\%$ with the final mask obtained by $\mathcal{L}_1$ and $\mathcal{L}_2$, while \textit{HyperSparse} leads to an intersection of approximately $35\%$.
This results show \textit{HyperSparse} changes less parameter while reordering weights, which implies that more structures from the dense model are exploited.
It also shows that \textit{HyperSparse} has a significantly smaller learning duration than $\mathcal{L}_1$ and $\mathcal{L}_2$.
The horizontal bars point to the intersection before last training epoch and show that $\mathcal{L}_1$ and $\mathcal{L}_2$ only intersect by $60\%$ and $50\%$, while \textit{HyperSparse} is getting very close to the final mask with more than $90\%$ intersection.
This indicates that \textit{HyperSparse} finds a more stable set of high valued weights and reduces exploration, as the mask has less variation in the final epochs.
More results for other training settings are shown in the supplementary material, Sec.~B.

Moreover, we analyze the resulting weight distribution of our method and compare it to \textit{IMP}~\cite{pmlr-v119-frankle20a} and \textit{SRatio}~\cite{NEURIPS2020_eae27d77}.
Fig.~\ref{fig:weight_distribution} shows the number of remaining weights per layer for ResNet-32 that consists of three scaling levels, which end up with the linear layer (LL).
Each scaling level consists of four residual blocks (RES), which are connected by a downsampling-block (DS).
The basic topology of \textit{ART} and \textit{IMP} looks similar, since both methods show a constant keep-ratio over the residual blocks.
Furthermore, \textit{ART} and \textit{IMP} use more parameters in downsampling and linear layers.
We conclude that these two layer types require more weights and consequently are more important to the model.
The higher accuracy discussed earlier suggest that our method exploit these weights better.
To show that this results are also obtained on other datasets, models, and sparsity levels, we describe further weight distributions in the supplementary material, Sec.~C and show that the number of parameters in the linear layer decreases drastically for a small set of classes in CIFAR-10.
Moreover, the compared method \textit{SRatio} assumes that suitable sparse networks can be obtained using handcrafted keep-ratios per layer.
It has a quadratic decreasing keep-ratio that can be observed in Fig.~\ref{fig:weight_distribution}. As shown in Tab.~\ref{table:prune_methods}, our method \textit{ART} performs significantly better than \textit{SRatio} and therefore we deduce that fixed keep-ratios have an adverse effect on performance.
Reordering weights during training favors well performing sparse NNs, especially in high sparsity regimes.

\subsection{What do networks compress first?}
\label{sec:whatDoNetworksCompressFirst}

Along with the introduction of \textit{ART}, we are faced with the question of which patterns are compressed first into the large weights that remain after magnitude pruning during regularization. This question is in contrast to Hooker's question \textit{``What Do Compressed Deep Neural Networks Forget?''}~\cite{https://doi.org/10.48550/arxiv.1911.05248} and challenges the fundamental assumption of magnitude pruning, which assumes large weights to be most important. In this section, we analyze the chronological order of how samples are compressed and introduce the metric \textit{Compression Position} (CP) to determine it. 

According to our method, regularization starts at epoch $e_S$ and ends at $e_E$ and therefore the weights $W$ have different states $\mathcal{W}=\{W_e\}_{e=e_S}^{e_E}$ during training. We measure the individual accuracy over time $\psi_\text{I}$ reached by the sparse network for a training sample $(x,y)\in S$, defined by 
\begin{equation}
\label{equ:sample_accuracy}
    \psi_\text{I}\big( x, y, f, \mathcal{W} \big) = \frac{\big|\big\{ W_e\in \mathcal{W}\ |\ f(\nu(W_e)\odot W_e, x) = y\big\}\big|} {e_E - e_S}.
\end{equation}
After computing the individual accuracy for all samples \mbox{$\Psi=\big\{\psi_\text{I}\big( x_n, y_n, f, \mathcal{W} \big)\big\}_{n=1}^N$} and sorting $\Psi$ in descending order, the metric $\text{CP}\big( x, y, f, \mathcal{W} \big)$ describes the relative position of $\psi_\text{I}\big( x, y, f, \mathcal{W} \big)$ in $\text{sort}(\Psi)$. In other words, early compressed and correctly classified samples obtain a low CP close to $0$, and those compressed later closer to $1$. 

We calculate the CP metric for all samples in CIFAR-10 during training of dense, low, and high sparsity NNs. The compression behaviour for dense NNs is measured during the pre-training phase ($e_S=0$ and $e_E=60$) and for sparse NNs during regularization phase ($e_S=60$ and $e_E=e_\text{max}$). 

To show, which samples are compressed first into the remaining highest weights, the $5\%$ samples with lowest CP are visualized in Fig.~\ref{fig:learning_clusters} in the latent space of the well known \textit{CLIP} framework~\cite{pmlr-v139-radford21a} mapped by \textit{t-SNE}~\cite{van2008visualizing}. As commonly known, the dense model compresses easy samples of all classes in the early stages~\cite{https://doi.org/10.48550/arxiv.2211.11355,NEURIPS2020_ea89621b}, while the low sparsity model already loses some. In the high sparsity regime no discriminative decision rules are left at beginning of training, and the remaining classes are compressed step by step as the training continues (see supplementary material, Sec.~E). In our experiments, we have seen continuously that there is a bias towards the class \textit{deer}. 
We call this effect \textit{``the deer bias''}, which must be reduced with regularisation. 
The \textit{deer} bias suggests that large weights in dense NNs do not encode decision rules for all classes.

To quantify the above results, Tab.~\ref{table:human_score_position} shows the average CP for all samples belonging to a specific class. Additionally, we split the class sets into four subsets according to their difficulty. We estimate the difficulty of a sample by counting the human label errors that are made from three human annotators derived from CIFAR-N~\cite{wei2022learning}, \eg, 2 means that two of three persons mislabeled the sample. The first observation is that the above mentioned separation of classes is confirmed, since CP values are similar in dense NNs, but diverge in sparse NNs. In high sparsity regimes, the \textit{deer} bias is persistent before first samples of other classes are compressed. The classes \textit{horse} and \textit{airplane} are only included at the end of the training. The second observation is, that within a closed set of samples belonging to a class, difficult samples are compressed later. This nature is similar to the training process of dense NNs.

Implementation details and more fine-grained results are available in the supplementary material, Sec.~E.
\section{Conclusion}

Our work presents \textit{Adaptive Regularized Training} (\textit{ART}), a method that utilizes regularization to obtain sparse neural networks. The regularization is amplified continuously and used to shrink most weight magnitudes close to zero. We introduce the novel regularization loss \textit{HyperSparse} that induces sparsity inherently while maintaining a well balanced tradeoff between exploration of new sparse topologies and exploitation of weights that remain after pruning. Extensive experiments on CIFAR and TinyImageNet show that our novel framework outperforms sparse learning competitors. \textit{HyperSparse} is superior to standard regularization losses and leads to impressive performance gains in extremely high sparsity regimes and is much faster. 
Additional investigations provide new insights about the weight distribution during network compression and about patterns that are encoded in high valued weights. 

Overall, this work provides new insights into sparse neural networks and helps to develop sustainable machine learning by reducing neural network complexity. 

\section{Acknowledgments}
This work was supported by the Federal Ministry of Education and Research (BMBF), Germany under the project AI service center KISSKI (grant no. 01IS22093C),
the Deutsche Forschungsgemeinschaft (DFG) under Germany’s Excellence Strategy within the Cluster of Excellence PhoenixD (EXC 2122),
and by the Federal Ministry of the Environment, Nature Conservation, Nuclear Safety and Consumer Protection, Germany under the project GreenAutoML4FAS (grant no. 67KI32007A).

{\small
\bibliographystyle{ieee_fullname}
\bibliography{egbib}
}

\renewcommand{\thesection}{\Alph{section}}
\setcounter{section}{0}%
\setcounter{subsection}{0}%
\setcounter{figure}{1}
\setcounter{table}{1}
\setcounter{equation}{1}
\setcounter{footnote}{0}%
\setcounter{page}{1}%
\clearpage
\section*{Supplementary Material}
\appendix
    
This document provides supplementary material for the paper  \textit{HyperSparse Neural Networks: Shifting Exploration to Exploitation through
Adaptive Regularization}.
At first, Sec.~\ref{appendix:experimentalSetup} gives detailed information about the implementation of our method.
Subsequently, Sec.~\ref{appendix:maskInterection} presents more detailed results of the intersection of largest weights during training and the final pruning mask. The weight distribution after training with our introduced method shown in the main paper is analyzed for a wider set of configurations in
Sec.~\ref{appendix:WeightDistribution}.
Moreover, 
Sec.~\ref{appendix:loss} and
Sec.~\ref{appendix:compression} elaborate the gradient and the compression behaviour during regularization presented in the main paper more into detail. 

\section{Detailed Experimental Setup}
\label{appendix:experimentalSetup}

As described in \cite{NEURIPS2020_eae27d77}, we evaluated our method on the datasets CIFAR-10/100~\cite{krizhevsky2009learning} and Tiny\-Image\-Net~\cite{deng2009imagenet} with the models ResNet-32~\cite{He_2016_CVPR} and VGG-19~\cite{simonyan2014very}.
CIFAR-10 is a dataset for a classification task with \num{50000} training and \num{10000} validation samples on 32x32 color-images labeled with 10 classes.
Respectively CIFAR-100 has 100 classes and the same amount of samples.
The dataset TinyImageNet consists of \num{100000} training and \num{10000} validation samples with an image-size of 64x64, where samples are labeled with a set of 200 classes.

As done in \cite{NEURIPS2020_eae27d77}, we train our models for 160 epochs on CIFAR-10/100 and 300 epochs on Tiny\-Image\-Net using SGD-optimizer, with an initial learning rate of 0.1 and a batch size of 64.
We decay the learning rate by factor 0.1 at epoch 2/4 and 3/4 of the total number of epochs.
The weight decay is set to $1\cdot10^{-4}$.
In our experiments all results are averaged over 5 runs.

In the original implementation of \textit{SmartRatio}~\cite{NEURIPS2020_eae27d77}, weights in the final linear layer are pruned with a fixed pruning rate of $70\%$.
Thus, too much weights remain when training on ResNet-32 with a pruning ratio of 99.8\% on dataset CIFAR-100 and TinyImageNet.
To this reason, we change the pruning ratio in the linear layer to $90\%$ for this two training settings only. 
The methods \textit{SNIP}~\cite{lee2018snip}, \textit{GraSP}~\cite{wang2019picking}, \textit{SmartRatio}~\cite{NEURIPS2020_eae27d77}, and \textit{LTH}~\cite{frankle2018lottery} suggest rules to obtain fixed masks. 
This mask is applied to the model weights before training.
In contrast, \textit{IMP}~\cite{pmlr-v119-frankle20a} iteratively trains a model to epoch $T$ and prunes $20\%$ of the remaining weights until the desired pruning rate is reached.
After each iteration the weights and learning rate are reset to epoch $k$ and retrained again to epoch $T$.
To be comparable, we define $k=20$ and $T=160$ for CIFAR-10/100 as well as $k=40$ and $T=300$ for TinyImageNet.
As described in \cite{pmlr-v119-evci20a}, RigL performes better with a longer training duration. 
To this reason we extend the optimization time of the uniform distributed RigL-method by training for 360 epochs with a learning rate of 0.1, followed by the fine-tuning-step of 160 epochs on CIFAR-10/100 and 300 epochs on Tiny\-Image\-Net. 
The fine-tuning step is equal to ART.
All further hyperparameters of RigL are adopted from \cite{pmlr-v119-evci20a}.

Our proposed method \textit{ART}, described in Sec.~3.2 in the main paper, consists of three steps.
In the first step we train our model to convergence for 60 epochs using a fix learning rate of $0.1$.
Subsequently we enable the used regularization term, with a small initialisation rate of $\lambda_{init}=5\cdot10^{-6}$ and increasing factor of $\eta=1.05$.
To reduce noise in choosing the best pruned model, we average the accuracy $\psi(\nu(W_e) \odot W_e)$ over epoch ($e-1$, $e$, $e+1$), where $e$ describes the current epoch and $\nu$ denotes magnitude pruning that obtains a binary mask.
The first two steps are used to obtain the weights and masks for fine-tuning. During fine-tuning, we use the training schedule described above as done in~\cite{NEURIPS2020_eae27d77}.

\section{Mask intersection in Regularized Training}
\label{appendix:maskInterection}

In this section we show further results of our experiments measuring the mask intersection over epoch~$e$ from Sec.~4.4 in the main paper.
We measure the relative overlap between the weights with highest magnitude at epoch $e$ and the final mask in different settings with different models, datasets, regularization losses, and pruning rates.
Therefore, Tab.~\ref{tab:keysMaskIntersec} shows the important keypoints of intersection at the end of pre-training (epoch 60) and one epoch before the final mask was found ($e=K-1$).
We observe that our regularization loss $\mathcal{L}_\text{HS}$ has a higher intersection in nearly all settings at epoch $60$ and epoch $K-1$ compared to $\mathcal{L}_1$ and $\mathcal{L}_2$ loss.
This indicates that our \textit{HyperSparse} loss changes less parameter while reordering weights from remaining to pruned and vice versa.

In addition, Tab.~\ref{tab:keysMaskIntersec} presents the total number of training epochs to obtain the final mask (including step~1 and step~2).
It shows that our \textit{HyperSparse} loss needs less epochs to terminate in nearly all settings.
Since \textit{ART} terminates, if the best pruned model outperforms the unpruned model at epoch $e$,
we deduce that $\mathcal{L}_\text{HS}$ creates a well performing sparse network faster compared to $\mathcal{L}_1$ and $\mathcal{L}_2$ loss.

\section{Weight Distribution}
\label{appendix:WeightDistribution}

In this section, we show further experiments of the weight distribution per layer in the final mask, as evaluated in Sec.~4.4 in the main paper.
We analyse the resulting masks for dataset CIFAR-10 and CIFAR-100, pruning-rate $\kappa \in \{90\%, 98\%, 99.5\%\}$ as well as for model ResNet-32 and VGG-19.
Weight distributions obtained by the methods \textit{IMP}~\cite{pmlr-v119-frankle20a}, \textit{SRatio}~\cite{NEURIPS2020_eae27d77} and \textit{ART} using \textit{HyperSpase} loss are analyzed. All values are averaged over 5 runs.

\begin{figure*}
\centering
\resizebox{0.95\textwidth}{!}{\begin{tikzpicture}
\begin{groupplot}[group style={group size= 2 by 4},
                    height = 0.2\textwidth,width=\columnwidth]
    \pgfplotsset{/pgfplots/group/.cd, vertical sep=0.5cm, horizontal sep=1.5cm}
    
   \nextgroupplot[
        title=CIFAR-10,
        width = 1\columnwidth,
        height = 0.2\linewidth,
        xmin = -2, xmax = 37,
        ymin = 0., ymax = 1.0,
        xticklabels=\empty,
        yticklabels=\empty,
        xtick = {1, 11, 14 ,22 ,25, 33},
        ytick={-10, 10}
    ]
    \draw[violet, thick] (0.9,0.1) rectangle (10.9,0.9) node[gray, pos=.5] {RES};
    
    \draw[teal, thick] (11.1,0.1) -- (13.9,0.3) -- (13.9,0.7) -- (11.1,0.9) -- cycle node[gray, anchor=west] at (11,.5) {\rotatebox{90}{DS}}; 
    
    \draw[violet, thick] (14.1,0.3) rectangle (21.9,0.7) node[gray, pos=.5] {RES};
    
    \draw[teal, thick] (22.1,0.3) -- (24.9,0.4) -- (24.9,0.6) -- (22.1,0.7) -- cycle node[gray, anchor=west] at (22,.5) {\rotatebox{90}{DS}}; 
    
    \draw[violet, thick] (25.1,0.4) rectangle (32.9,0.6) node[gray, pos=.5] {RES};
    
    \draw[purple, thick] (33.1,0.1) rectangle (34.9,0.9) node[gray, pos=.5] {\rotatebox{90}{LL}};  

    \nextgroupplot[
        title=CIFAR-100,
        width = 1\columnwidth,
        height = 0.2\linewidth,
        xmin = -2, xmax = 37,
        ymin = 0., ymax = 1.0,
        xticklabels=\empty,
        yticklabels=\empty,
        xtick = {1, 11, 14 ,22 ,25, 33},
        ytick={-10, 10}
    ]
    \draw[violet, thick] (0.9,0.1) rectangle (10.9,0.9) node[gray, pos=.5] {RES};
    
    \draw[teal, thick] (11.1,0.1) -- (13.9,0.3) -- (13.9,0.7) -- (11.1,0.9) -- cycle node[gray, anchor=west] at (11,.5) {\rotatebox{90}{DS}}; 
    
    \draw[violet, thick] (14.1,0.3) rectangle (21.9,0.7) node[gray, pos=.5] {RES};
    
    \draw[teal, thick] (22.1,0.3) -- (24.9,0.4) -- (24.9,0.6) -- (22.1,0.7) -- cycle node[gray, anchor=west] at (22,.5) {\rotatebox{90}{DS}}; 
    
    \draw[violet, thick] (25.1,0.4) rectangle (32.9,0.6) node[gray, pos=.5] {RES};
    
    \draw[purple, thick] (33.1,0.1) rectangle (34.9,0.9) node[gray, pos=.5] {\rotatebox{90}{LL}};  

    \nextgroupplot[
        bar width=3pt,
        scaled ticks=false, 
        log ticks with fixed point,
        tick label style={/pgf/number format/fixed},
        xtick = {1, 11, 14 ,22 ,25, 33},
        xmin = -2, xmax = 37,
        ymin = 0, ymax = 30000,
        grid = both,
        minor tick num = 0,
        major grid style = {lightgray},
        minor grid style = {lightgray!50},
        width = 1\columnwidth,
        height = 0.35\linewidth,
        ylabel = {$\kappa=90\%$},
    ]
    
    \addplot[red, const plot, thick] table [x=l_numb,y=p_keep_mean, col sep=comma] {ablation_results/param_per_layer/PpL_method=IMP_dataset=cifar10_model=resnet32_pruneRate=0.9.txt};
    \addplot[blue, const plot, thick] table [x=l_numb,y=p_keep_mean, col sep=comma] {ablation_results/param_per_layer/PpL_method=smartRatio_dataset=cifar10_model=resnet32_pruneRate=0.9.txt};
    \addplot[OliveGreen, const plot, thick] table [x=l_numb,y=p_keep_mean, col sep=comma] {ablation_results/param_per_layer/PpL_method=RegularMask_dataset=cifar10_model=resnet32_pruneRate=0.9.txt};
    
    \coordinate (top) at (rel axis cs:0,1);
    \nextgroupplot[
        bar width=3pt,
        scaled ticks=false, 
        log ticks with fixed point,
        tick label style={/pgf/number format/fixed},
        xtick = {1, 11, 14 ,22 ,25, 33},
        xmin = -2, xmax = 37,
        ymin = 0, ymax = 30000,
        grid = both,
        minor tick num = 0,
        major grid style = {lightgray},
        minor grid style = {lightgray!50},
        width = 1\columnwidth,
        height = 0.35\linewidth,
    ]
    
    \addplot[red, const plot, thick] table [x=l_numb,y=p_keep_mean, col sep=comma] {ablation_results/param_per_layer/PpL_method=IMP_dataset=cifar100_model=resnet32_pruneRate=0.9.txt};
    \addplot[blue, const plot, thick] table [x=l_numb,y=p_keep_mean, col sep=comma] {ablation_results/param_per_layer/PpL_method=smartRatio_dataset=cifar100_model=resnet32_pruneRate=0.9.txt};
    \addplot[OliveGreen, const plot, thick] table [x=l_numb,y=p_keep_mean, col sep=comma] {ablation_results/param_per_layer/PpL_method=RegularMask_dataset=cifar100_model=resnet32_pruneRate=0.9.txt};

    \nextgroupplot[
        bar width=3pt,
        scaled ticks=false, 
        log ticks with fixed point,
        tick label style={/pgf/number format/fixed},
        xtick = {1, 11, 14 ,22 ,25, 33},
        xmin = -2, xmax = 37,
        ymin = 0, ymax = 7000,
        grid = both,
        minor tick num = 0,
        major grid style = {lightgray},
        minor grid style = {lightgray!50},
        width = 1\columnwidth,
        height = 0.35\linewidth,
        ylabel = {$\kappa=98\%$},
    ]
    
    \addplot[red, const plot, thick] table [x=l_numb,y=p_keep_mean, col sep=comma] {ablation_results/param_per_layer/PpL_method=IMP_dataset=cifar10_model=resnet32_pruneRate=0.98.txt};
    \addplot[blue, const plot, thick] table [x=l_numb,y=p_keep_mean, col sep=comma] {ablation_results/param_per_layer/PpL_method=smartRatio_dataset=cifar10_model=resnet32_pruneRate=0.98.txt};
    \addplot[OliveGreen, const plot, thick] table [x=l_numb,y=p_keep_mean, col sep=comma] {ablation_results/param_per_layer/PpL_method=RegularMask_dataset=cifar10_model=resnet32_pruneRate=0.98.txt};

    \nextgroupplot[
        bar width=3pt,
        scaled ticks=false, 
        log ticks with fixed point,
        tick label style={/pgf/number format/fixed},
        xtick = {1, 11, 14 ,22 ,25, 33},
        xmin = -2, xmax = 37,
        ymin = 0, ymax = 7000,
        grid = both,
        minor tick num = 0,
        major grid style = {lightgray},
        minor grid style = {lightgray!50},
        width = 1\columnwidth,
        height = 0.35\linewidth,
    ]
    
    \addplot[red, const plot, thick] table [x=l_numb,y=p_keep_mean, col sep=comma] {ablation_results/param_per_layer/PpL_method=IMP_dataset=cifar100_model=resnet32_pruneRate=0.98.txt};
    \addplot[blue, const plot, thick] table [x=l_numb,y=p_keep_mean, col sep=comma] {ablation_results/param_per_layer/PpL_method=smartRatio_dataset=cifar100_model=resnet32_pruneRate=0.98.txt};
    \addplot[OliveGreen, const plot, thick] table [x=l_numb,y=p_keep_mean, col sep=comma] {ablation_results/param_per_layer/PpL_method=RegularMask_dataset=cifar100_model=resnet32_pruneRate=0.98.txt};

    \nextgroupplot[
        bar width=3pt,
        scaled ticks=false, 
        log ticks with fixed point,
        tick label style={/pgf/number format/fixed},
        xtick = {1, 11, 14 ,22 ,25, 33},
        xmin = -2, xmax = 37,
        ymin = 0, ymax = 5000,
        grid = both,
        minor tick num = 0,
        major grid style = {lightgray},
        minor grid style = {lightgray!50},
        width = 1\columnwidth,
        height = 0.35\linewidth,
        xlabel = {Layer Index $i$},
        ylabel = {$\kappa=99.5\%$},
    ]
    
    \addplot[red, const plot, thick] table [x=l_numb,y=p_keep_mean, col sep=comma] {ablation_results/param_per_layer/PpL_method=IMP_dataset=cifar10_model=resnet32_pruneRate=0.995.txt};
    \addplot[blue, const plot, thick] table [x=l_numb,y=p_keep_mean, col sep=comma] {ablation_results/param_per_layer/PpL_method=smartRatio_dataset=cifar10_model=resnet32_pruneRate=0.995.txt};
    \addplot[OliveGreen, const plot, thick] table [x=l_numb,y=p_keep_mean, col sep=comma] {ablation_results/param_per_layer/PpL_method=RegularMask_dataset=cifar10_model=resnet32_pruneRate=0.995.txt};

    \nextgroupplot[
        bar width=3pt,
        scaled ticks=false, 
        log ticks with fixed point,
        tick label style={/pgf/number format/fixed},
        xtick = {1, 11, 14 ,22 ,25, 33},
        xmin = -2, xmax = 37,
        ymin = 0, ymax = 5000,
        grid = both,
        minor tick num = 0,
        major grid style = {lightgray},
        minor grid style = {lightgray!50},
        width = 1\columnwidth,
        height = 0.35\linewidth,
        xlabel = {Layer Index $i$},
        legend style = {
          at={(-0.05,-0.3)},
          anchor=north,
          legend columns=5},
    ]
    
    \addplot[red, const plot, thick] table [x=l_numb,y=p_keep_mean, col sep=comma] {ablation_results/param_per_layer/PpL_method=IMP_dataset=cifar100_model=resnet32_pruneRate=0.995.txt};
    \addplot[blue, const plot, thick] table [x=l_numb,y=p_keep_mean, col sep=comma] {ablation_results/param_per_layer/PpL_method=smartRatio_dataset=cifar100_model=resnet32_pruneRate=0.995.txt};
    \addplot[OliveGreen, const plot, thick] table [x=l_numb,y=p_keep_mean, col sep=comma] {ablation_results/param_per_layer/PpL_method=RegularMask_dataset=cifar100_model=resnet32_pruneRate=0.995.txt};

    \coordinate (bot) at (rel axis cs:1,0);
    
    \legend{\textit{IMP}, \textit{SRatio}, \textit{ART} + $\mathcal{L}_\text{HS}$}

\end{groupplot}

\path (top-|current bounding box.west) -- node[anchor=south,rotate=90] {Number of Weights per Layer after Pruning with Pruning Rate~$\kappa$} (bot-|current bounding box.west);

\end{tikzpicture}}
\vspace{-0pt}
\caption{Distribution of weights per layer in \textbf{ResNet-32} after pruning. We visualize the results in the left column for CIFAR-10 and right column for CIFAR-100 as well as for pruning rates $\kappa \in \{90\%, 98\%, 99.5\%\}$ in each row.
The layer index describes the execution order which means that higher indices are calculated later in inference.
All results are averaged over 5 runs.
We group the model in residual blocks (RES), downsampling blocks (DS) and the linear layer (LL).
Our Method \textit{ART} + $\mathcal{L}_\text{HS}$ distributes weights comparable to \textit{IMP}~\cite{pmlr-v119-frankle20a}, but has more weights in the downsampling layers.
The method \textit{SRatio}~\cite{NEURIPS2020_eae27d77} has a quadratic decreasing keep-ratio.
We observe the linear layer in CIFAR-100 deserves more weights compared to CIFAR-10, due to the bigger number of classes.
}
\label{fig:weight_distribution_appendig_resnet}
\vspace{0pt}
\end{figure*}
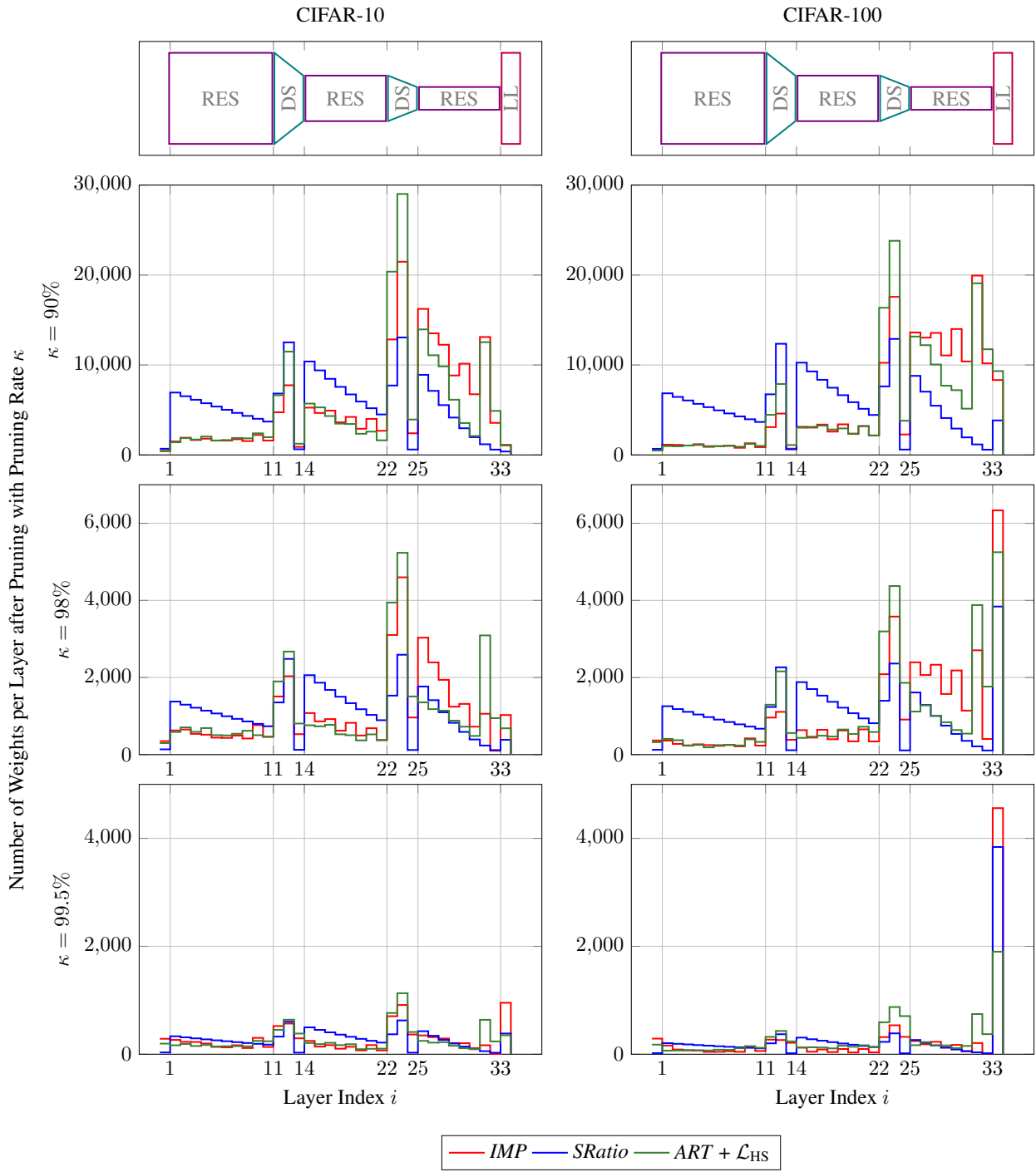

In Fig.~\ref{fig:weight_distribution_appendig_resnet}, we show the resulting weight distributions for ResNet-32.
Note that the model is grouped in three residual blocks (RES), two downsampling blocks (DS) and a linear layer (LL).
We observe that \textit{ART} + $\mathcal{L}_\text{HS}$ and \textit{IMP} have comparable distributions of weights.
Both methods show a relative constant distribution in the residual layers, except the last one.
This last layer has an decreasing number of weights, especially in the simpler task given in CIFAR-10.
In comparison, \textit{SRatio} uses a fixed keep-ratio in a quadratic decreasing manner and thus the weight distribution is not dependent on data.
Since \textit{ART} and \textit{IMP} outperform \textit{SRatio} by far in accuracy (Tab.~1 in the main paper), this hand-crafted rule has adverse effect on performance.
Moreover, we observe a relatively high number of weights in the downsampling layer for \textit{ART} and \textit{IMP}, which indicates that these layers are more important.

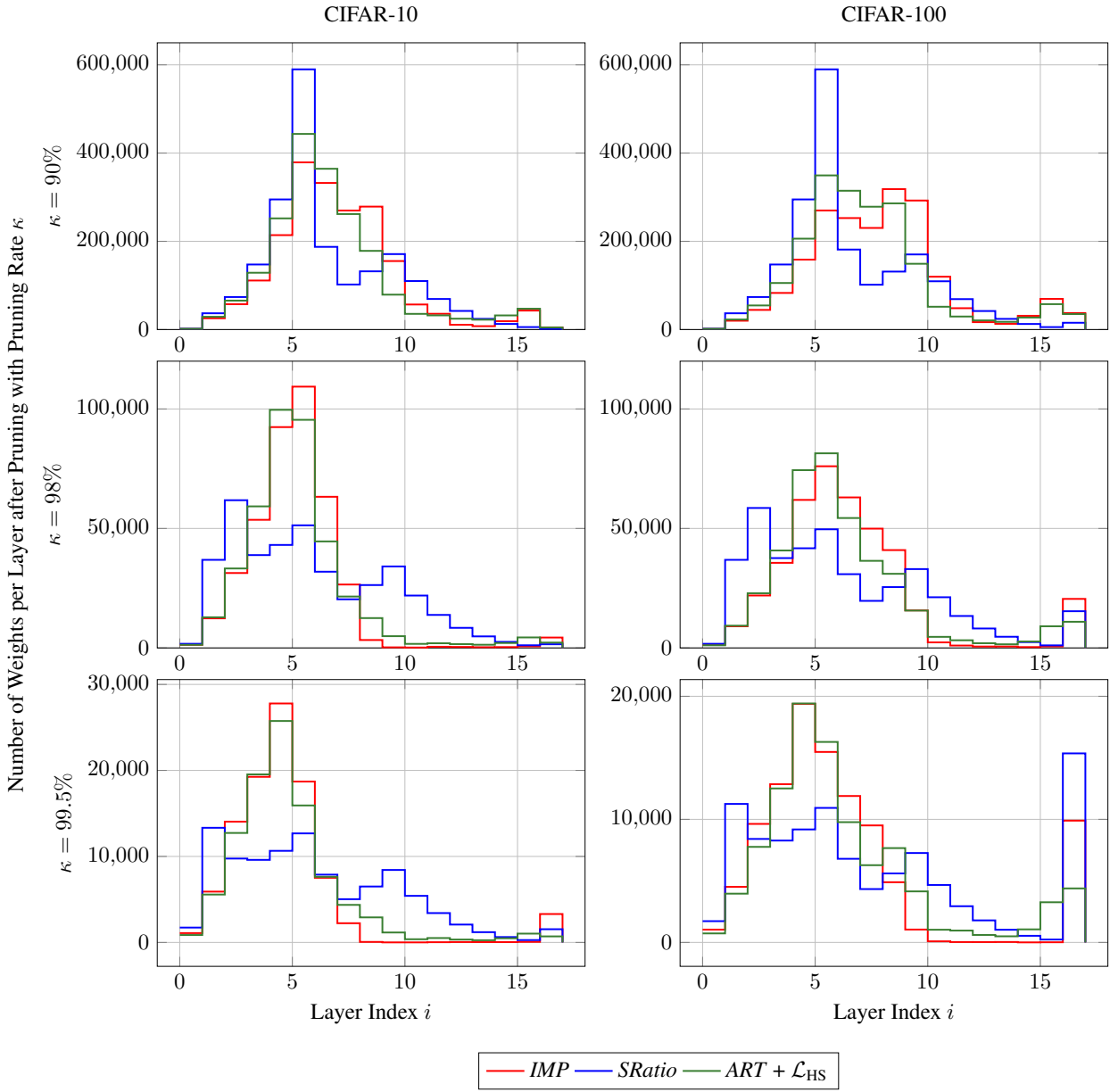
\begin{figure*}
\centering
\resizebox{0.95\textwidth}{!}{\begin{tikzpicture}
\begin{groupplot}[group style={group size= 2 by 3},
                    height = 0.2\linewidth,width=\columnwidth]
    \pgfplotsset{/pgfplots/group/.cd, vertical sep=0.5cm, horizontal sep=1.5cm}

    \nextgroupplot[
        title={CIFAR-10},
        bar width=3pt,
        scaled ticks=false, 
        log ticks with fixed point,
        tick label style={/pgf/number format/fixed},
        xmin = -1, xmax = 18,
        ymin = 0, ymax = 650000,
        grid = both,
        minor tick num = 0,
        major grid style = {lightgray},
        minor grid style = {lightgray!50},
        width = 1\columnwidth,
        height = 0.35\linewidth,
        ylabel = {$\kappa=90\%$},
    ]
    
    \addplot[red, const plot, thick] table [x=l_numb,y=p_keep_mean, col sep=comma] {ablation_results/param_per_layer/PpL_method=IMP_dataset=cifar10_model=vgg19_pruneRate=0.9.txt};
    \addplot[blue, const plot, thick] table [x=l_numb,y=p_keep_mean, col sep=comma] {ablation_results/param_per_layer/PpL_method=smartRatio_dataset=cifar10_model=vgg19_pruneRate=0.9.txt};
    \addplot[OliveGreen, const plot, thick] table [x=l_numb,y=p_keep_mean, col sep=comma] {ablation_results/param_per_layer/PpL_method=RegularMask_dataset=cifar10_model=vgg19_pruneRate=0.9.txt};
    
    \coordinate (topl) at (rel axis cs:0,1);

    \nextgroupplot[
        title={CIFAR-100},
        bar width=3pt,
        scaled ticks=false, 
        log ticks with fixed point,
        tick label style={/pgf/number format/fixed},
        xmin = -1, xmax = 18,
        ymin = 0, ymax = 650000,
        grid = both,
        minor tick num = 0,
        major grid style = {lightgray},
        minor grid style = {lightgray!50},
        width = 1\columnwidth,
        height = 0.35\linewidth,
    ]
    
    \addplot[red, const plot, thick] table [x=l_numb,y=p_keep_mean, col sep=comma] {ablation_results/param_per_layer/PpL_method=IMP_dataset=cifar100_model=vgg19_pruneRate=0.9.txt};
    \addplot[blue, const plot, thick] table [x=l_numb,y=p_keep_mean, col sep=comma] {ablation_results/param_per_layer/PpL_method=smartRatio_dataset=cifar100_model=vgg19_pruneRate=0.9.txt};
    \addplot[OliveGreen, const plot, thick] table [x=l_numb,y=p_keep_mean, col sep=comma] {ablation_results/param_per_layer/PpL_method=RegularMask_dataset=cifar100_model=vgg19_pruneRate=0.9.txt};
    
    \coordinate (topr) at (rel axis cs:1,1);
    
    \nextgroupplot[
        bar width=3pt,
        scaled ticks=false, 
        log ticks with fixed point,
        tick label style={/pgf/number format/fixed},
        xmin = -1, xmax = 18,
        ymin = 0, ymax = 120000,
        grid = both,
        minor tick num = 0,
        major grid style = {lightgray},
        minor grid style = {lightgray!50},
        width = 1\columnwidth,
        height = 0.35\linewidth,
        ylabel = {$\kappa=98\%$},
    ]
    
    \addplot[red, const plot, thick] table [x=l_numb,y=p_keep_mean, col sep=comma] {ablation_results/param_per_layer/PpL_method=IMP_dataset=cifar10_model=vgg19_pruneRate=0.98.txt};
    \addplot[blue, const plot, thick] table [x=l_numb,y=p_keep_mean, col sep=comma] {ablation_results/param_per_layer/PpL_method=smartRatio_dataset=cifar10_model=vgg19_pruneRate=0.98.txt};
    \addplot[OliveGreen, const plot, thick] table [x=l_numb,y=p_keep_mean, col sep=comma] {ablation_results/param_per_layer/PpL_method=RegularMask_dataset=cifar10_model=vgg19_pruneRate=0.98.txt};

    \nextgroupplot[
        bar width=3pt,
        scaled ticks=false, 
        log ticks with fixed point,
        tick label style={/pgf/number format/fixed},
        xmin = -1, xmax = 18,
        ymin = 0, ymax = 120000,
        grid = both,
        minor tick num = 0,
        major grid style = {lightgray},
        minor grid style = {lightgray!50},
        width = 1\columnwidth,
        height = 0.35\linewidth,
    ]
    
    \addplot[red, const plot, thick] table [x=l_numb,y=p_keep_mean, col sep=comma] {ablation_results/param_per_layer/PpL_method=IMP_dataset=cifar100_model=vgg19_pruneRate=0.98.txt};
    \addplot[blue, const plot, thick] table [x=l_numb,y=p_keep_mean, col sep=comma] {ablation_results/param_per_layer/PpL_method=smartRatio_dataset=cifar100_model=vgg19_pruneRate=0.98.txt};
    \addplot[OliveGreen, const plot, thick] table [x=l_numb,y=p_keep_mean, col sep=comma] {ablation_results/param_per_layer/PpL_method=RegularMask_dataset=cifar100_model=vgg19_pruneRate=0.98.txt};

    \nextgroupplot[
        bar width=3pt,
        scaled ticks=false, 
        log ticks with fixed point,
        tick label style={/pgf/number format/fixed},
        xmin = -1, xmax = 18,
        grid = both,
        minor tick num = 0,
        major grid style = {lightgray},
        minor grid style = {lightgray!50},
        width = 1\columnwidth,
        height = 0.35\linewidth,
        xlabel = {Layer Index $i$},
        ylabel = {$\kappa=99.5\%$},
    ]
    
    \addplot[red, const plot, thick] table [x=l_numb,y=p_keep_mean, col sep=comma] {ablation_results/param_per_layer/PpL_method=IMP_dataset=cifar10_model=vgg19_pruneRate=0.995.txt};
    \addplot[blue, const plot, thick] table [x=l_numb,y=p_keep_mean, col sep=comma] {ablation_results/param_per_layer/PpL_method=smartRatio_dataset=cifar10_model=vgg19_pruneRate=0.995.txt};
    \addplot[OliveGreen, const plot, thick] table [x=l_numb,y=p_keep_mean, col sep=comma] {ablation_results/param_per_layer/PpL_method=RegularMask_dataset=cifar10_model=vgg19_pruneRate=0.995.txt};

    \nextgroupplot[
        bar width=3pt,
        scaled ticks=false, 
        log ticks with fixed point,
        tick label style={/pgf/number format/fixed},
        xmin = -1, xmax = 18,
        grid = both,
        minor tick num = 0,
        major grid style = {lightgray},
        minor grid style = {lightgray!50},
        width = 1\columnwidth,
        height = 0.35\linewidth,
        xlabel = {Layer Index $i$},
        legend style = {
          at={(-0.05,-0.3)},
          anchor=north,
          legend columns=5},
    ]
    
    \addplot[red, const plot, thick] table [x=l_numb,y=p_keep_mean, col sep=comma] {ablation_results/param_per_layer/PpL_method=IMP_dataset=cifar100_model=vgg19_pruneRate=0.995.txt};
    \addplot[blue, const plot, thick] table [x=l_numb,y=p_keep_mean, col sep=comma] {ablation_results/param_per_layer/PpL_method=smartRatio_dataset=cifar100_model=vgg19_pruneRate=0.995.txt};
    \addplot[OliveGreen, const plot, thick] table [x=l_numb,y=p_keep_mean, col sep=comma] {ablation_results/param_per_layer/PpL_method=RegularMask_dataset=cifar100_model=vgg19_pruneRate=0.995.txt};
    
    \coordinate (bot) at (rel axis cs:1,0);
    
    \legend{\textit{IMP}, \textit{SRatio}, \textit{ART} + $\mathcal{L}_\text{HS}$}

\end{groupplot}
\path (topl-|current bounding box.west) -- node[anchor=south,rotate=90] {Number of Weights per Layer after Pruning with Pruning Rate~$\kappa$} (bot-|current bounding box.west);
\end{tikzpicture}}
\vspace{0pt}
\caption{Distribution of weights per layer in \textbf{VGG-19} after pruning. We visualize the results in the left column for CIFAR-10 and right column for CIFAR-100 as well as for pruning rates $\kappa \in \{90\%, 98\%, 99.5\%\}$ in each row.
The layer index describes the execution order which means that higher indices are calculated later in inference.
All results are averaged over 5 runs.
In \cite{NEURIPS2020_eae27d77}, VGG-19 is constructed using multiple convolutional layers, which irregularly increase  in the number of parameter over index $i$.
The model ends with a linear layer.
Our Method \textit{ART} + $\mathcal{L}_\text{HS}$ distributes weights comparable to \textit{IMP}~\cite{pmlr-v119-frankle20a}, while SRatio~\cite{NEURIPS2020_eae27d77} uses more weights in layers with higher index.
We observe the linear layer in CIFAR-100 deserves more weights compared to CIFAR-10, due to the bigger number of classes.
}
\label{fig:weight_distribution_appendig_vgg}
\vspace{0pt}
\end{figure*}

Further, we present the weight distribution for VGG-19 in Fig.~\ref{fig:weight_distribution_appendig_vgg}.
We observe that the layer around index 5 has more weights for \textit{ART} and \textit{IMP}.
Nearly no weights remain in layer with index higher than 10, except the final linear layer.
Considering the increasing sparsity, the weight distribution is shifted towards the earlier layers with low index.
We deduce that in higher sparsity regimes the weight in earlier layer are more important in VGG-19.
The handcrafted rule of \textit{SRatio} shows a relatively flat weight distribution.
Overall, the number of weights in the linear layer increases for CIFAR-100, due to the increasing number of classes compared to CIFAR-10 in ResNet-32 and for VGG-19.

\section{HyperSparse Gradient Analysis}
\label{appendix:loss}
In this section, we analyze the gradient of our \textit{HyperSparse} regularization loss with respect to the model weights $w\in W$. Assuming that important weights have large magnitudes, we show that \textit{HyperSparse} subsides to no regularisation for important values and evolves to a strong penalization for unimportant values. This behaviour allows exploitation in the set of the important weights that remain after magnitude pruning. 
Furthermore, we show that our loss ensures a smooth transition in the gradient between unimportant and important weights, such that exploration in the set of unimportant weights is possible during training.  

\paragraph{Gradient.} Our loss evolves sparseness and adapts on the weight magnitude by utilizing the non-linearity of the \textit{Hyper}bolic Tangent function
\begin{equation}
    \label{equ:tanh}
    \tanh(x) = \frac{e^x - e^{-x}}{e^z + e^{-x}} \quad \in (-1,1),
\end{equation}
which is the reason for the name \textit{HyperSparse}.
The maximum of the derivative of $\tanh$ is $1$ at \mbox{$x=0$} and strongly vanishes close to zero for large values:
\begin{equation}
    \label{equ:tanh_properties}
    \argmax_x \frac{d \tanh(x)}{d{x}} = 0 \quad \text{and} \quad \lim\limits_{x \rightarrow \pm \infty}{\frac{d \tanh(x)}{d{x}}}=0 \ .
\end{equation}
In this paper, the Hyperbolic Tangent function of a magnitude $\tanh(|\cdot |)$ is denoted by $t(\cdot)$ for simplicity. 

For the sake of completeness, we recapitulate the definition of \textit{HyperSparse} from Eq.~(2) in the main paper:
\begin{equation}
\label{equ:spartan_appendix}
\begin{split}
    \mathcal{L}_\text{HS}(W) = & \frac{1}{A} \sum_{i=1}^{|W|} \bigg( |w_i| 
    \sum_{j=1}^{|W|} t(s\cdot w_j)\bigg) - \sum_{i=1}^{|W|} |w_i|    
     \\
     & \text{with}\quad  A:= \sum_{w\in W} t(s\cdot w) \\
     & \text{and} \quad \forall w\in W:\quad \frac{dA}{dw} = 0,
\end{split}
\end{equation}

and want to note again, that $A$ denotes a pseudo-constant term that is considered to be a constant in the gradient computation, and $s$ is a scaling factor discussed in the end of this section. 
In this section, sum notations as $\sum_{i=1}^{|W|}$ will be simplified by $\sum_{w_i}$ or by $\sum_{w}$ if $w$ is unique.
Furthermore, we will leave out declarations of set memberships like $w_i\in W$ and state that every $w$ is in the set of model weights $W$. Also the scope of formulations is consistently defined as $\forall w\in W$. 

With this notations and simplifications, the derivative of Eq.\eqref{equ:spartan_appendix} w.r.t.~to a weight $w_i$ can be defined as follows:
\begin{equation}
\label{equ:spartan_grad1_appendix}
\begin{split}
    \frac{d\mathcal{L}_{HS}}{dw_i} = & 
     \underbrace{\frac{d}{dw_i} \frac{|w_i| \sum_{w_j}t(s\cdot w_j)}{A}}_{\textbf{I}} \ \ + \\  & \quad \underbrace{\frac{d}{dw_i} \frac{\sum_{w_n\neq w_i} |w_n| \sum_{w_j}t(s\cdot w_j)}{A}}_{\textbf{II}} - \text{sign}(w_i) 
     \\
    = & 
       \text{sign}(w_i) \cdot \underbrace{\frac{|w_i|\cdot t'(s\cdot w_i) + \sum_{w_j}t(s\cdot w_j)}{A}}_{\textbf{I}}  \ \ + 
     \\  & \quad  \text{sign}(w_i) \cdot \underbrace{\frac{\sum_{w_n\neq w_i} |w_n| \cdot t'(s\cdot w_i)}{A}}_{\textbf{II}} - \text{sign}(w_i)
    \\
    = & 
    \text{sign}(w_i) \cdot \bigg[
       \frac{\sum_{w_j}t(s\cdot w_j)}{A={\sum_{w_j} t(s\cdot w_j)}} \ \ +  \\ &\quad \frac{\sum_{w_n} |w_n| \cdot t'(s\cdot w_i)}{A={\sum_{w_j} t(s\cdot w_j)}} - 1 \bigg]
    \\
    = & 
       \text{sign}(w_i) \cdot  \frac{t'(s\cdot w_i) \cdot \sum_{w_n} |w_n| }{{\sum_{w_j} t(s\cdot w_j)}} \quad .
\end{split}
\end{equation}
The gradient consists of a term that is depending on the weight distribution in $W$ and the derivative $t'=\frac{dt}{dw_i}$ at the considered weights magnitude $|w_i|$ scaled with $s$. The behaviour of \textit{HyperSparse} can be explained with the gradients for very small and very large magnitudes: For large magnitudes $|w_i|\gg0$, the derivative in Eq.~\eqref{equ:spartan_grad1_appendix} collapses to 
\begin{equation}
\label{equ:spartan_grad2_appendix}
\frac{d\mathcal{L}_{HS}}{dw_i} \bigg|_{|s\cdot w_i|\gg0}\approx \text{sign}(w_i) \cdot 0 = 0
\end{equation}
which is effectively no regularisation. 
For very small values $w_i\approx0$, the derivative 
\begin{equation}
\label{equ:spartan_grad3_appendix}
\frac{d\mathcal{L}_{HS}}{dw_i} \bigg|_{|s\cdot w_i|\approx0}\approx \text{sign}(w_i) \cdot \frac{\sum_{w_n} |w_n| }{{\sum_{w_j} t(s\cdot w_j)}} 
\end{equation}
is larger and increases, if the weights in $W$ are clearly separated in two sets of important (large magnitude) and unimportant weights (low magnitude). The gradient of weights that are not assigned to one of those sets is between Eq.~\eqref{equ:spartan_grad2_appendix} and~\eqref{equ:spartan_grad3_appendix} and therefore allows an easier exploration of those weights during training. 

\paragraph{Aligning with $s$.}
\label{appendix:gradient_s}
In the definition of \textit{HyperSparse}, the scaling factor $s$ aligns the loss with the actual weight distribution. The aim is that weights $|w|>|w_\kappa|$ are not or softly and $|w|<|w_\kappa|$ strongly penalized. A weight distribution does not need to be aligned with the derivative of the Hyperbolic Tangent function such that large weights are mapped close to $0$ and small weights close to $1$. To fix this, we align $W$ by scaling $w_\kappa$ with $s$ so that it lies on the inflection point of the gradient. The desired scaling factor can be derived by
\begin{equation}
    t'''(s\cdot |w_\kappa|\approx 0.6585) = 0
\end{equation}
and setting $s=\frac{0.6586}{|w_\kappa|}$. Large scaling factors $s$ lead to rampant gradient distribution at weight $w_\kappa$ towards weights of low magnitude. Examples can be found in Fig.~1 
in the main paper.

\begin{table}
    \centering
    \setlength{\tabcolsep}{3pt}
    \resizebox{\linewidth}{!}{

    \begin{tabular}{ccc|ccc|ccc|ccc}%
    \toprule
    & & &   \multicolumn{3}{c}{\shortstack{\textbf{\#Epochs} \\ \textbf{to Final Mask}}} & 
            \multicolumn{3}{c}{\shortstack{\textbf{Intersection} \\ \textbf{at} $\boldsymbol{e=60}$}} & 
            \multicolumn{3}{c}{\shortstack{\textbf{Intersection} \\ \textbf{at} $\boldsymbol{e=K-1}$ }}\\
     &   & \shortstack{$\boldsymbol{\kappa}$} & $\mathcal{L}_\text{HS}$ & $\mathcal{L}_1$ & $\mathcal{L}_2$ & $\mathcal{L}_\text{HS}$ & $\mathcal{L}_1$ & $\mathcal{L}_2$ & $\mathcal{L}_\text{HS}$ & $\mathcal{L}_1$ & $\mathcal{L}_2$ \\
    
    \midrule
    
    \multirow{6}{*}{\hfil \rotatebox{90}{\shortstack{\textbf{CIFAR-10}}} \hfil}
     & \multirow{3}{*}{\hfil \rotatebox{90}{\shortstack{\textbf{ResNet-}\\\textbf{32}}} \hfil}
       & $90\%$ & 87 & 95  & 23 & 0.46 & 0.35 & 0.41 & 0.84 & 0.72 & 0.83  \\
     & & $98\%$ & 112 & 130 & 194 & 0.29 & 0.15 & 0.1 & 0.86 & 0.54 & 0.4  \\
     & & $99.5\%$ & 136 & 152 & 177 & 0.23 & 0.12 & 0.17 & 0.91 & 0.54 & 0.49  \\
     
    \cmidrule(rr){2-12}
     & \multirow{3}{*}{\hfil \rotatebox{90}{\shortstack{\textbf{VGG-}\\\textbf{19}}} \hfil}
       & $90\%$ & 64 & 66 & 185 & 0.77 & 0.71 & 0.46 & 0.87 & 0.83 & 0.67  \\
     & & $98\%$ & 76 & 81 & 230 & 0.51 & 0.45 & 0.02 & 0.83 & 0.76 & 0.7  \\
     & & $99.5\%$ & 104 & 118 & 232 & 0.32 & 0.19 & 0.02 & 0.82 & 0.59 & 0.52  \\
     
    \midrule
    \multirow{6}{*}{\hfil \rotatebox{90}{\shortstack{\textbf{CIFAR-100}}} \hfil}

     & \multirow{3}{*}{\hfil \rotatebox{90}{\shortstack{\textbf{VGG-}\\\textbf{19}}} \hfil} 
       & $90\%$ & 68 & 71 & 211 & 0.66 & 0.6 & 0.13 & 0.85 & 0.81 & 0.67  \\
     & & $98\%$ & 96 & 106 & 238 & 0.39 & 0.28 & 0.02 & 0.83 & 0.64 & 0.66  \\
     & & $99.5\%$ & 116 & 121 & 226 & 0.27 & 0.16 & 0.01 & 0.85 & 0.62 & 0.48  \\
    
    \cmidrule(rr){2-12}
     & \multirow{3}{*}{\hfil \rotatebox{90}{\shortstack{\textbf{ResNet-}\\\textbf{32}}} \hfil}
       & $90\%$ & 99 & 113 & 177 & 0.45 & 0.28 & 0.2 & 0.88 & 0.63 & 0.49  \\
     & & $98\%$ & 123 & 134 & 183 & 0.34 & 0.2 & 0.18 & 0.91 & 0.61 & 0.49  \\
     & & $99.5\%$ & 147 & 150 & 163 & 0.26 & 0.19 & 0.28 & 0.95 & 0.7 & 0.62  \\
    
    \bottomrule
    
    \end{tabular}
    }

    \caption{Complementary key-points to the experiments about mask intersection in Sec.~4.4 in the main paper.
    The intersection indicates the relative overlap of weights with highest magnitude during training to remaining weights in the final mask obtained by \textit{ART}.
    For simplicity, we only show the intersection at the end of pre-training ($e=60$) and one epoch before the final mask was found ($e=K-1$).
    In addition, this table shows the number of training epochs to the final mask.
    The information are demonstrated for $\mathcal{L}_1$, $\mathcal{L}_2$ and \textit{HyperSparse} loss $\mathcal{L}_{HS}$, over different pruning rates $\kappa$, models and datasets.
    }
    \label{tab:keysMaskIntersec}
\end{table}

\section{Interpretable Compression}
\label{appendix:compression}

This chapter discusses the process of knowledge compression that compresses patterns from a pre-trained dense network into a sparse network that consists of the set of $1-\kappa$ highest weights, where $\kappa$ denotes the desired pruning rate. 

The first subsection presents the CIFAR-N~\cite{wei2022learning} dataset that is used to analyze the compression behavior in Sec.~4.5. We show how it relates to CIFAR~\cite{krizhevsky2009learning} and how we visualize the label distribution with the modern \textit{CLIP} framework~\cite{pmlr-v139-radford21a}. Then we elaborate the introduced metric \textit{Compression Position} more in detail. For the sake of completeness, we lastly present and discuss figures and tables that show results for additional settings that could not be presented in the main paper due to lack of space.

\subsubsection*{CIFAR-N}
To analyze human-like label errors and to provide real-world label noise for researchers, \textit{Wei}~\etal introduced the CIFAR-N~\cite{wei2022learning} dataset that uses the CIFAR~\cite{krizhevsky2009learning} training data $S=\{(x_n, y_n)\}^N_{n=1}$, but has different ground truth labels. Every sample was labeled by $3$ different persons, inducing their subjective human bias, such that the dataset formulation can be defined as $S=\{(x_n, \{y_n^1,y_n^2,y_n^3\})\}^N_{n=1}$. They show, that single persons consistently induce an error rate between 10-20\% (compared to original CIFAR). Moreover, they show that human-like label noise is harder to tackle in robust learning scenarios compared to synthetic label noise.

\begin{figure*}
     \centering
     \begin{subfigure}[b]{0.47\textwidth}
         \centering
         \includegraphics[width=\textwidth]{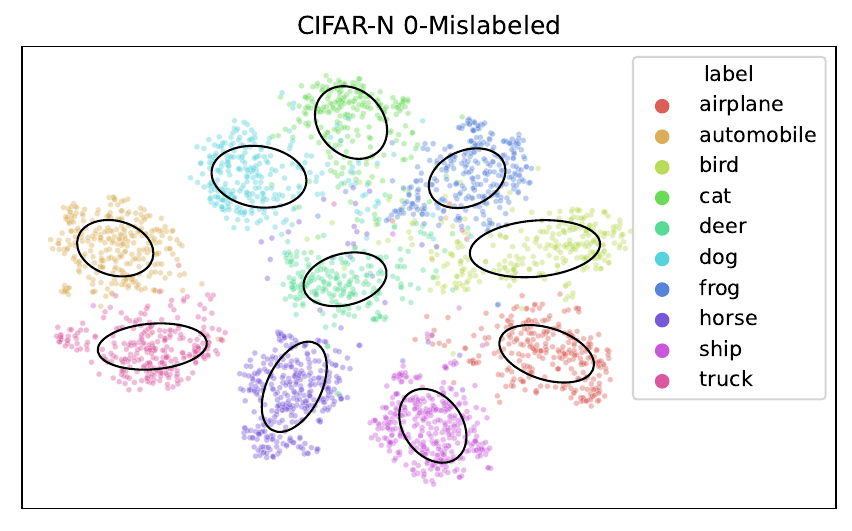}
     \end{subfigure}
     \begin{subfigure}[b]{0.47\textwidth}
         \centering
         \includegraphics[width=\textwidth]{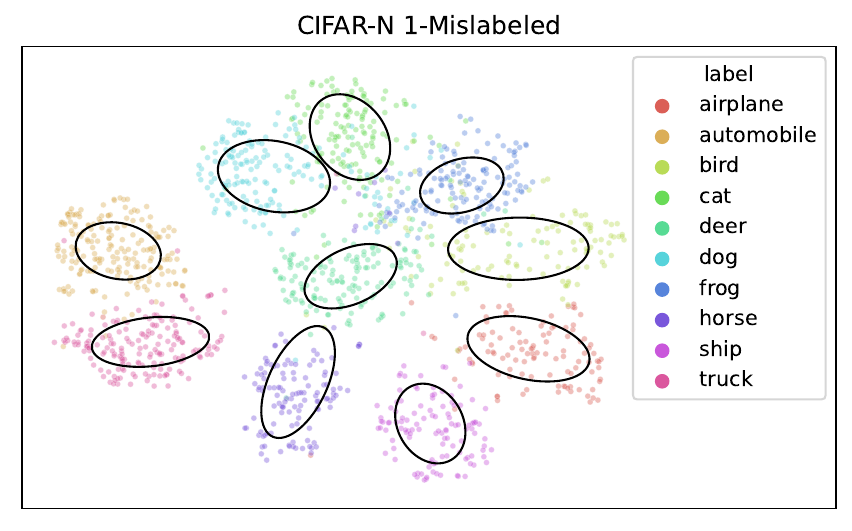}
     \end{subfigure}
     \begin{subfigure}[b]{0.47\textwidth}
         \centering
         \includegraphics[width=\textwidth]{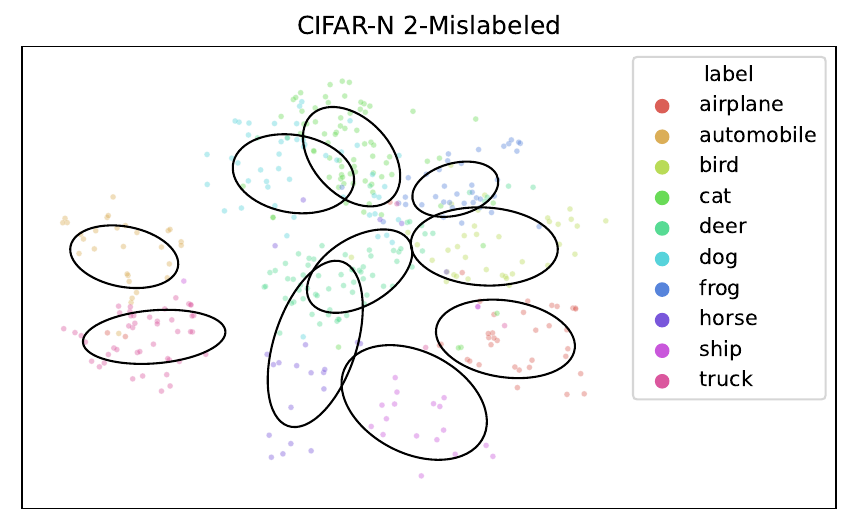}
     \end{subfigure}
     \begin{subfigure}[b]{0.47\textwidth}
         \centering
         \includegraphics[width=\textwidth]{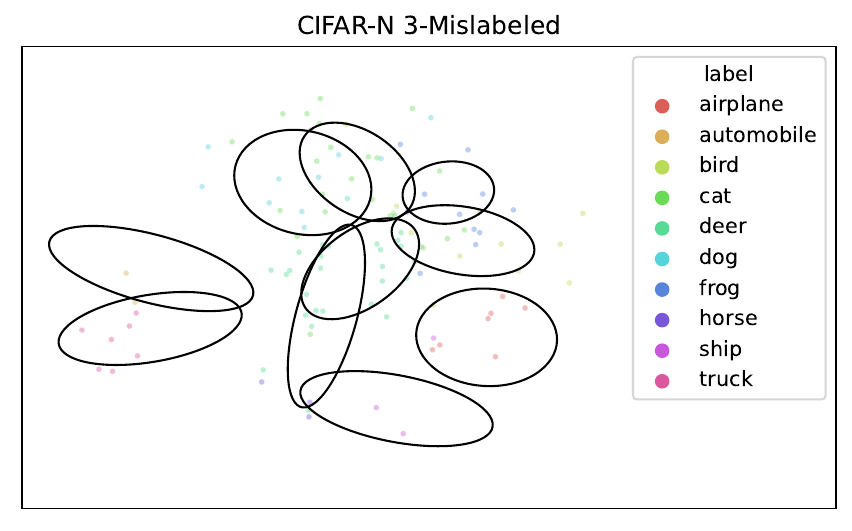}
     \end{subfigure}
        \caption{CIFAR-10N samples in the \textit{CLIP} latent space, mapped by \textit{t-SNE} into two dimensions. The datset is split into four subsets deduced by the hardness score explained in Sec.~\ref{appendix:compression}. The term ``$h$-Mislabeled'' explains that $h$ of 3 persons mislabeled the corresponding sample. Ellipses indicate the double standard deviation of a class in the \textit{t-SNE} space. It shows that samples with higher hardness score $h$ lead to a larger standard deviation and suggest that \textit{CLIP} in combination with \textit{t-SNE} is a suitable visualization tool to show visualize human-like recognition behaviour.}
        \label{fig:appendix_cifarN}
\end{figure*}

We use the multi-label $y^m_i$ from CIFAR-N and the most likely correct label $y_i$ from CIFAR-10 and derive a ``hardness-score'' $h_i$. For a sample $(x_i, y_i)$, the score 
\begin{equation}
h_i = \big|\big\{ y_i^m \in\{y_i^1, y_i^2, y_i^3\}\big\} | y_i^m \neq y_i\big\}\big| \quad \in [0,3]
\end{equation}
describes, how often a sample was mislabeled in CIFAR-N and therefore relates to the difficulty. To illustrate the distribution of classes and labels, we map all images of CIFAR-10 to the latent space of the high performing diffusion model~\textit{CLIP}~\cite{pmlr-v139-radford21a} that is using vision transformers~\cite{dosovitskiy2021an} and is trained on large training data. After mapping the images to the \textit{CLIP} latent space, we reduce the dimensions with the \textit{t-SNE}~\cite{van2008visualizing} algorithm to two dimensions as shown in Fig.~\ref{fig:appendix_cifarN}. The four sub-figures split the samples from CIFAR-10 according to their score $h_i$. It shows that all classes have samples with every score. Moreover, the variance of the samples per class grows with increasing hardness. As harder samples are more likely to have a larger distance to cluster centers, because they differ to the ``easy'' and unambiguous class templates, the increasing variance indicates that the \textit{CLIP} latent space combined with \textit{t-SNE} is a good tool to visualize a human-like sample distribution. 

According to the well known and often discussed effect that samples with easy patterns and unambiguous labels are memorized first~\cite{https://doi.org/10.48550/arxiv.2211.11355,NEURIPS2020_ea89621b, pmlr-v70-arpit17a}, samples with a higher hardness-score should be compressed later in the training process. We use the hardness-score to evaluate if this effect is also present in the process of compressing patterns from a pre-trained dense neural network into a dense sub-network using our method.

\subsubsection*{Compression Position (CP)}
The next section formally defines the evaluation metric \textit{Compression Position} as described in Sec.~4.5 in the main paper.
Measurement of classification capabilities of neural networks $f(W, x)$ is usually performed by the accuracy metric 
\begin{equation}
\label{equ:appendix_accuracy}
    \psi\big( S, f, W \big) = \frac{\big|\big\{(x, y)\in S\ |\ f(W, x) = y\big\}\big|} {|S|}.
\end{equation}
The accuracy of a specific class can be obtained by calculating $\psi$ for a subset $S_c\subset S$ with only samples of a specific class $y=c$. To answer the question \textit{``Which classes are represented first in a neural network?''}, one can measure the class accuracy after every training epoch $e$ and plot them. To reduce the complex plot into a single metric, the area-under-curve (AUC) could be obtained for every specific class. Drawbacks from the AUC mesurements are, that the absolute values of AUC are not comparable between different settings (\textit{i.e.}, datasets, models, \ldots). For example, large and complex data will lead to lower AUCs. Moreover, the class specific accuracy metric is not satisfying for the question \textit{``Which classes are compressed first into the higher magnitude weights?''} that is addressed in this paper. We noticed, that the class accuracy of sparse networks underlie high noise rates and are therefore hard to interpret. 

To tackle the drawbacks and generate a suitable metric for our work, we introduce the \textit{Compression Position} (CP) metric that is basically a sample based accuracy over time. It aims to quantify the relative position in time between epoch $e_S$ and $e_E$, where a sample $x$ is compressed from the dense weights $W$ into the weights with high magnitude $\nu(W)$, so that the sparse neural network is able to predict the correct ground truth label $f(\nu(W)\odot W,x)=y$.  

First, we redefine Eq.~\eqref{equ:appendix_accuracy} into a individual sample based accuracy for the pruned model that is defined as
\begin{equation}
\label{equ:appendix_sample_accuracy}
    \psi_\text{I}\big( x, y, f, \mathcal{W} \big) = \frac{\big|\big\{W_e\in \mathcal{W}\ |\ f(\nu(W_e)\odot W_e, x) = y\big\}\big|} {e_E - e_S}
\end{equation}
and $\mathcal{W}=\{W_e\}_{e=e_S}^{e_E}$ denotes a set of weight sets during the training between epoch $e_S$ and $e_E$.
The CP metric of $x$ is the normalized position of $\psi_\text{I}\big( x, y, f, \mathcal{W} \big)$ in a sorted list of the sample accuracy $\Psi=\text{sort}\bigg(\big\{ \psi_\text{I}\big( x_n, y_n, f, \mathcal{W} \big)\big\}_{n=1}^N\bigg)$ in descending order, such that 
\begin{equation}
    \text{CP}(x, y, f, \mathcal{W}, S): \  \psi_\text{I}(x, y, f, \mathcal{W}) \overset{!}{=} \Psi_{{\text{CP}}(x, y,f, \mathcal{W}) \cdot |\Phi|}
\end{equation}
holds.
The CP metric indicates the temporal position when a sample is compressed into the sparse weights $\nu(W)\odot W$, because CP increases if the corresponding sample is classified correct early and continuously in the training process.

\subsubsection*{Compression Behaviour}
\begin{figure*}
    \centering
    \begin{subfigure}[c]{\textwidth}
    \centering
     \begin{subfigure}[c]{0.28\textwidth}
         \centering
         \includegraphics[trim={0.3cm 0 0.2cm 0},clip, width=\textwidth]{figures/dense/0.pdf}
     \end{subfigure}
     \begin{subfigure}[c]{0.28\textwidth}
         \centering
         \includegraphics[trim={0.3cm 0 0.2cm 0},clip,width=\textwidth]{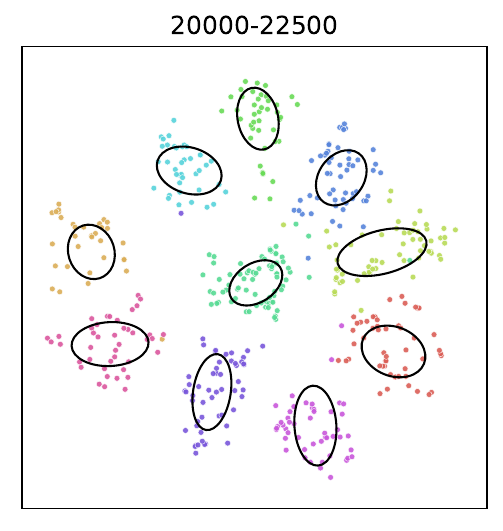}
     \end{subfigure}
     \begin{subfigure}[c]{0.28\textwidth}
         \centering
         \includegraphics[trim={0.3cm 0 0.2cm 0},clip,width=\textwidth]{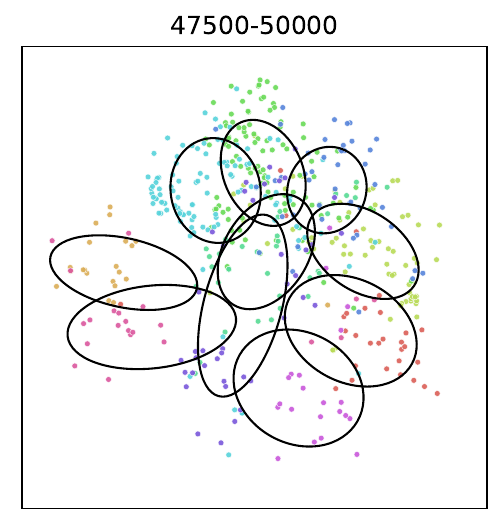}
     \end{subfigure}
     \hfill
     \begin{subfigure}[c]{0.13\textwidth}
         \centering         
         \vspace{5pt}
         \includegraphics[width=\textwidth]{figures/legend.pdf}
     \end{subfigure}
    \caption{Dense Model ($\kappa=0\%$)}
    \end{subfigure}
    
    \begin{subfigure}[c]{\textwidth}
    \centering
     \begin{subfigure}[c]{0.28\textwidth}
         \centering
         \includegraphics[trim={0.3cm 0 0.2cm 0},clip, width=\textwidth]{figures/0.9/0.pdf}
     \end{subfigure}
     \begin{subfigure}[c]{0.28\textwidth}
         \centering
         \includegraphics[trim={0.3cm 0 0.2cm 0},clip,width=\textwidth]{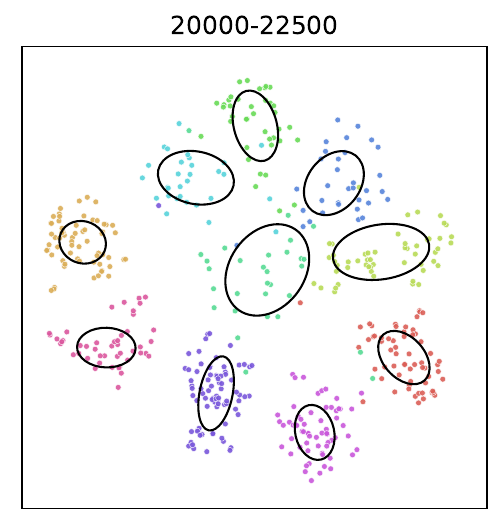}
     \end{subfigure}
     \begin{subfigure}[c]{0.28\textwidth}
         \centering
         \includegraphics[trim={0.3cm 0 0.2cm 0},clip,width=\textwidth]{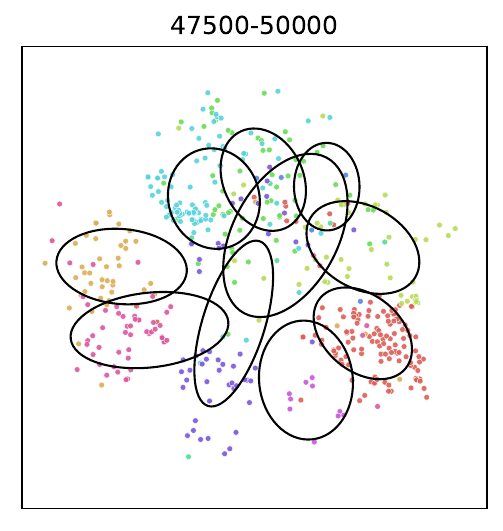}
     \end{subfigure}
     \hfill
     \begin{subfigure}[c]{0.13\textwidth}
         \centering
         \vspace{5pt}
         \includegraphics[width=\textwidth]{figures/legend.pdf}
     \end{subfigure}
    \caption{Low Sparsity ($\kappa=90\%$)}
    \end{subfigure}
    
        \begin{subfigure}[c]{\textwidth}
    \centering
     \begin{subfigure}[c]{0.28\textwidth}
         \centering
         \includegraphics[trim={0.3cm 0 0.2cm 0},clip, width=\textwidth]{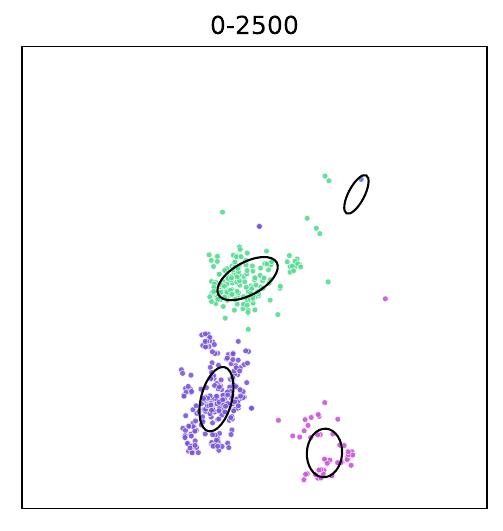}
     \end{subfigure}
     \begin{subfigure}[c]{0.28\textwidth}
         \centering
         \includegraphics[trim={0.3cm 0 0.2cm 0},clip,width=\textwidth]{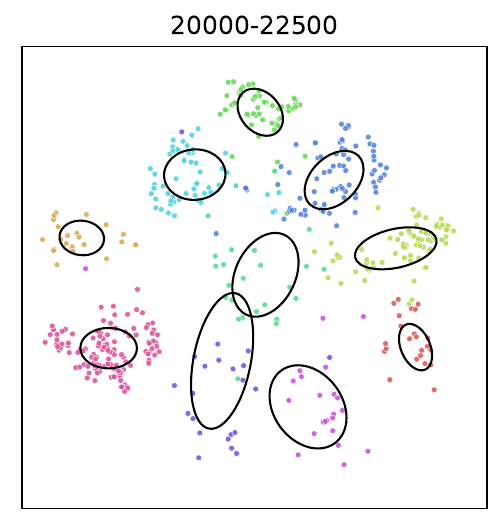}
     \end{subfigure}
     \begin{subfigure}[c]{0.28\textwidth}
         \centering
         \includegraphics[trim={0.3cm 0 0.2cm 0},clip,width=\textwidth]{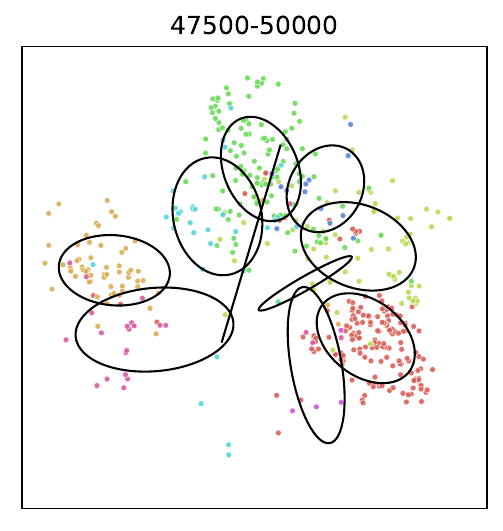}
     \end{subfigure}
     \hfill
     \begin{subfigure}[c]{0.13\textwidth}
         \vspace{5pt}
         \centering
         \includegraphics[width=\textwidth]{figures/legend.pdf}
     \end{subfigure}
    \caption{Low Sparsity ($\kappa=95\%$)}
    \end{subfigure}
    \caption{First 5\%, 40\%-45\%, and 95\%-100\% CIFAR-10 samples that are compressed into the remaining highest weights after pruning with $\kappa \in \{0\%, 90\%, 95\%\}$ deduced by the CP-metric. While dense networks learn samples approximately uniform-distributed over classes, the highest weights compress decision rules only for a subset of classes in the early learning stage. Note that we sampled by factor 10 for visualization purposes and ellipses represent the double standard deviation of cluster centers.}
    \label{fig:appendix_learning_cluster1}
\end{figure*}

\begin{figure*}
    \centering
    \begin{subfigure}[c]{\textwidth}
    \centering
     \begin{subfigure}[c]{0.28\textwidth}
         \centering
         \includegraphics[trim={0.3cm 0 0.2cm 0},clip, width=\textwidth]{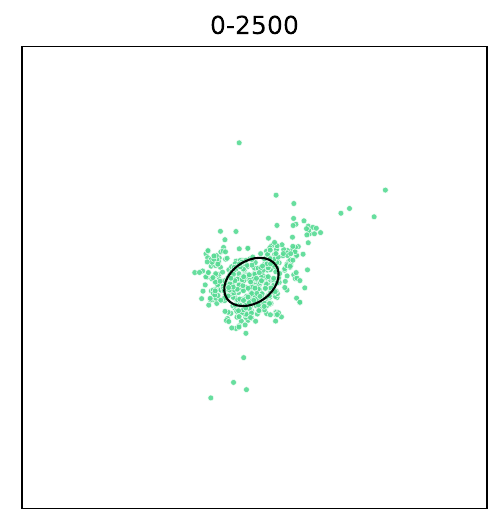}
     \end{subfigure}
     \begin{subfigure}[c]{0.28\textwidth}
         \centering
         \includegraphics[trim={0.3cm 0 0.2cm 0},clip,width=\textwidth]{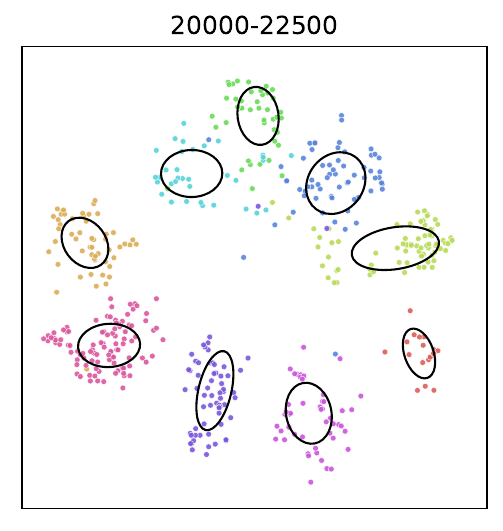}
     \end{subfigure}
     \begin{subfigure}[c]{0.28\textwidth}
         \centering
         \includegraphics[trim={0.3cm 0 0.2cm 0},clip,width=\textwidth]{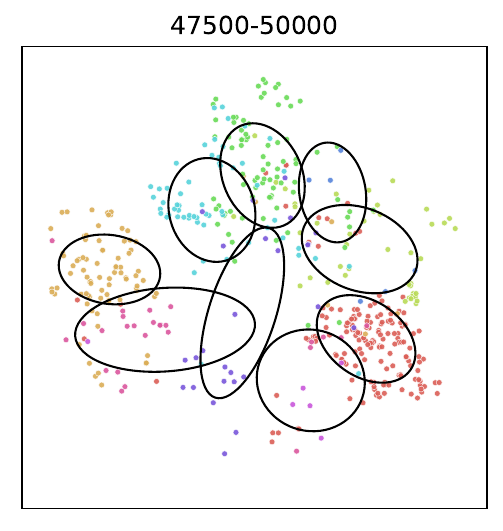}
     \end{subfigure}
     \hfill
     \begin{subfigure}[c]{0.13\textwidth}
         \centering         
         \vspace{5pt}
         \includegraphics[width=\textwidth]{figures/legend.pdf}
     \end{subfigure}
    \caption{Low Sparsity ($\kappa=98\%$)}
    \end{subfigure}
    
    \begin{subfigure}[c]{\textwidth}
    \centering
     \begin{subfigure}[c]{0.28\textwidth}
         \centering
         \includegraphics[trim={0.3cm 0 0.2cm 0},clip, width=\textwidth]{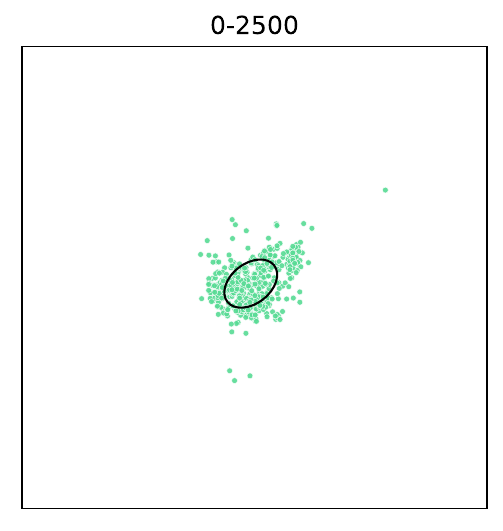}
     \end{subfigure}
     \begin{subfigure}[c]{0.28\textwidth}
         \centering
         \includegraphics[trim={0.3cm 0 0.2cm 0},clip,width=\textwidth]{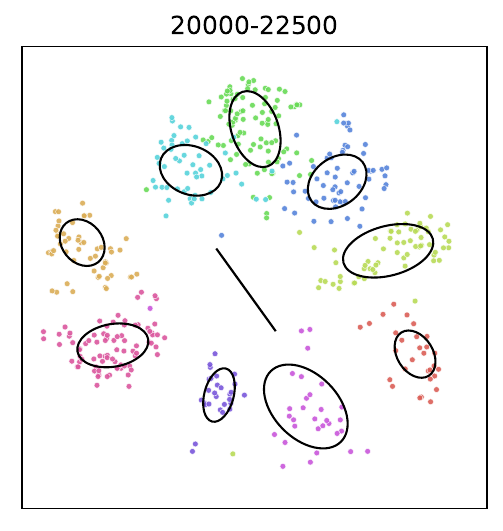}
     \end{subfigure}
     \begin{subfigure}[c]{0.28\textwidth}
         \centering
         \includegraphics[trim={0.3cm 0 0.2cm 0},clip,width=\textwidth]{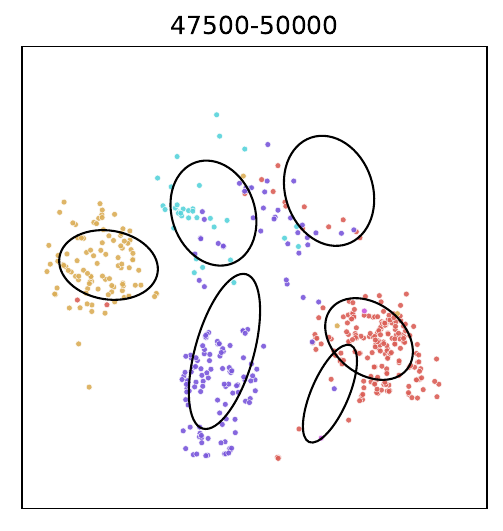}
     \end{subfigure}
     \hfill
     \begin{subfigure}[c]{0.13\textwidth}
         \centering
         \vspace{5pt}
         \includegraphics[width=\textwidth]{figures/legend.pdf}
     \end{subfigure}
    \caption{High Sparsity ($\kappa=99.5\%$)}
    \end{subfigure}
    
        \begin{subfigure}[c]{\textwidth}
    \centering
     \begin{subfigure}[c]{0.28\textwidth}
         \centering
         \includegraphics[trim={0.3cm 0 0.2cm 0},clip, width=\textwidth]{figures/0.998/0.pdf}
     \end{subfigure}
     \begin{subfigure}[c]{0.28\textwidth}
         \centering
         \includegraphics[trim={0.3cm 0 0.2cm 0},clip,width=\textwidth]{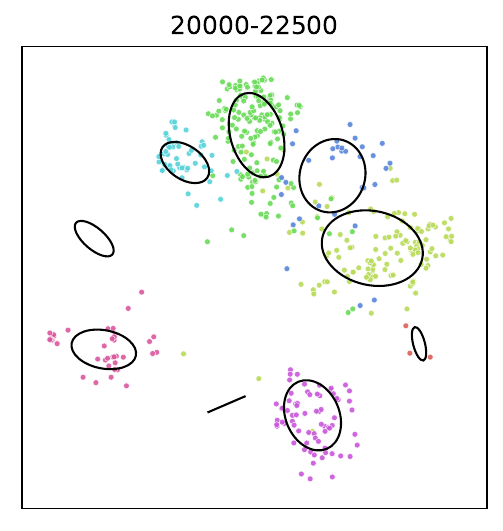}
     \end{subfigure}
     \begin{subfigure}[c]{0.28\textwidth}
         \centering
         \includegraphics[trim={0.3cm 0 0.2cm 0},clip,width=\textwidth]{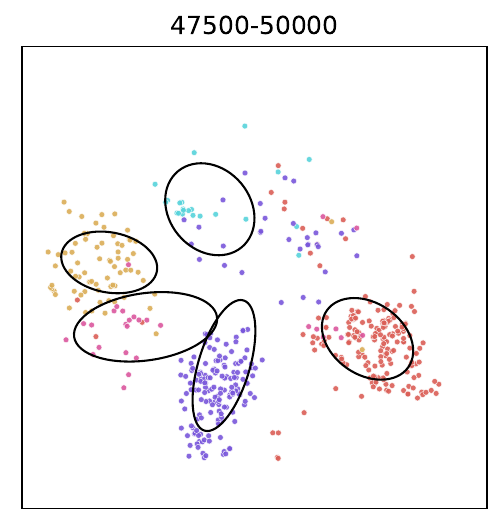}
     \end{subfigure}
     \hfill
     \begin{subfigure}[c]{0.13\textwidth}
         \vspace{5pt}
         \centering
         \includegraphics[width=\textwidth]{figures/legend.pdf}
     \end{subfigure}
    \caption{High Sparsity ($\kappa=99.8\%$)}
    \label{fig:appendix_learning_cluster2_c}
    \end{subfigure}
    \caption{First 5\%, 40\%-45\%, and 95\%-100\% CIFAR-10 samples that are compressed into the remaining highest weights after pruning with $\kappa \in \{98\%, 99.5\%, 99.8\%\}$ deduced by the CP-metric. While dense networks learn samples approximately uniform-distributed over classes, the highest weights compress decision rules only for a subset of classes in the early learning stage. Note that we sampled by factor 10 for visualization purposes and ellipses represent the double standard deviation of cluster centers.}
    \label{fig:appendix_learning_cluster2}
\end{figure*}

We present the main impressions of our investigations about the compression behaviour on class level in Tab.~3 and on sample level in Fig.~4 in the main paper. For the sake of completeness and to strengthen the claims, we report more detailed results in Tab.~\ref{table:appendix_human_score_position} and Fig.~\ref{fig:appendix_learning_cluster1} and~\ref{fig:appendix_learning_cluster2}.  

The order of compression $\Psi_\text{sort}$ for a dense, two low sparsity, and three high sparsity networks is visualized in Fig.~\ref{fig:appendix_learning_cluster1} and~\ref{fig:appendix_learning_cluster2}. Every sub-figure shows a consecutive set of \num{2500} samples from $\Psi_\text{sort}$ and gives an intuition, which patterns are compressed into the sparse network in the beginning, middle phase and end of training. 
First, we observe that the diversity of classes in the first \num{2500} samples decreases with increasing sparsity. Second, it shows that the intra-class variance increases over time. 
The first observation suggests that the highest weights do not make any decisions at the beginning, or only between a few classes. In the same way that only a few classes are compressed at the beginning, the remaining classes are compressed in isolation at the end (see Fig.~\ref{fig:appendix_learning_cluster2_c}). This is important for magnitude pruning based methods and high sparsity rates: If the highest weights have no capabilities in classification for all classes after dense training, perhaps the basic assumption that highest weights encode most important decision rules is wrong. Interestingly, our experiments consistently show, that the class \textit{deer} tends to be compressed first and moreover, \textit{deer} is the center in the \textit{t-SNE} mapped latent space of \textit{CLIP}. It seems like \textit{deer} is the general prototype of the dataset and therefore we call the effect of preferring one class in the first compression stage \textit{The deer bias}.
The second observation reveals the main commonality between dense training and compression through regularization. Derived from the human ability to reproduce simple patterns faster, dense and sparse networks learn the general patterns first during compression and encode the high frequency samples later.  

The Tab.~\ref{table:appendix_human_score_position} quantifies the results discussed before. It shows the compression rate for every class in CIFAR-10, subdivided by the hardness score introduced earlier. The dense networks compression rate for every class is more or less uniform-distributed. This promotes the first observation that all classes are encoded into the weights at the same time in dense networks. With increasing pruning rate $\kappa$, the classes are successively compressed into the high weights during regularization. The second observation is confirmed by dividing the classes according to their human label errors. The samples with higher label error are consistently compressed later into the high weights.

\begin{table*}
\centering
\small
\resizebox{\linewidth}{!}{
\begin{tabular}{ccc|cccccccccc} 
 \toprule
 \multirow{2}{*}{\textbf{CP}} & & \multicolumn{1}{c}{\multirow{2}{*}{\shortstack[c]{\textbf{Human} \\ \textbf{Label} \\ \textbf{Errors}}}} & \multicolumn{10}{c}{\textbf{CIFAR-10 Class}} \\ \cmidrule{4-13}

 &&  \multicolumn{1}{c}{}&
 ~~airplane &
 automobile &
 bird~ &
 cat~ &
 deer~ &
 dog~ &
 frog~ &
 horse~ &
 ship~ &
 truck~~ \\
 \midrule
 
\multirow{30}{*}{
    \rotatebox{90}{\shortstack[c]{\textbf{Sparsity Level}\\ \textbf{($\boldsymbol{\kappa}$)}}}}

&\multirow{4}{*}{\rotatebox{90}{\shortstack[c]{\textbf{Dense}\\ \textbf{Model} \\ \textbf{(0\%)}}}} & 0 & 0.452 & 0.295 & 0.554 & 0.639 & 0.482 & 0.551 & 0.425 & 0.429 & 0.357 & 0.400  \\
&& 1 & 0.547 & 0.329 & 0.632 & 0.722 & 0.548 & 0.628 & 0.482 & 0.493 & 0.439 & 0.479  \\
&& 2 & 0.676 & 0.389 & 0.741 & 0.808 & 0.653 & 0.699 & 0.572 & 0.710 & 0.592 & 0.588  \\
&& 3 & 0.783 & 0.542 & 0.823 & 0.862 & 0.760 & 0.826 & 0.665 & 0.769 & 0.720 & 0.695  \\
\cmidrule{2-13}
 
 &\multirow{4}{*}{\rotatebox{90}{\shortstack[c]{\textbf{Low}\\ \textbf{Sparsity} \\ \textbf{(90\%)}}}} & 0 & 0.580 & 0.612 & 0.499 & 0.494 & 0.166 & 0.460 & 0.257 & 0.524 & 0.470 & 0.567  \\
&& 1 & 0.671 & 0.640 & 0.573 & 0.601 & 0.217 & 0.540 & 0.303 & 0.570 & 0.522 & 0.613  \\
&& 2 & 0.790 & 0.666 & 0.678 & 0.721 & 0.292 & 0.625 & 0.376 & 0.737 & 0.631 & 0.676  \\
&& 3 & 0.841 & 0.764 & 0.772 & 0.796 & 0.392 & 0.754 & 0.461 & 0.808 & 0.762 & 0.710  \\
\cmidrule{2-13}
 
  &\multirow{4}{*}{\rotatebox{90}{\shortstack[c]{\textbf{Low}\\ \textbf{Sparsity} \\ \textbf{(95\%)}}}} & 0 & 0.739 & 0.685 & 0.604 & 0.696 & 0.187 & 0.486 & 0.400 & 0.173 & 0.223 & 0.543  \\
&& 1 & 0.796 & 0.709 & 0.665 & 0.763 & 0.217 & 0.538 & 0.439 & 0.212 & 0.253 & 0.578  \\
&& 2 & 0.868 & 0.733 & 0.745 & 0.836 & 0.280 & 0.581 & 0.517 & 0.383 & 0.363 & 0.625  \\
&& 3 & 0.912 & 0.808 & 0.822 & 0.880 & 0.366 & 0.701 & 0.574 & 0.468 & 0.494 & 0.655  \\
\cmidrule{2-13}
 
  &\multirow{4}{*}{\rotatebox{90}{\shortstack[c]{\textbf{Low}\\ \textbf{Sparsity} \\ \textbf{(98\%)}}}} & 0 & 0.742 & 0.635 & 0.579 & 0.640 & 0.045 & 0.504 & 0.349 & 0.471 & 0.313 & 0.452  \\
&& 1 & 0.798 & 0.666 & 0.643 & 0.721 & 0.053 & 0.563 & 0.395 & 0.523 & 0.350 & 0.492  \\
&& 2 & 0.870 & 0.696 & 0.727 & 0.800 & 0.066 & 0.623 & 0.479 & 0.700 & 0.452 & 0.546  \\
&& 3 & 0.906 & 0.797 & 0.797 & 0.851 & 0.079 & 0.753 & 0.547 & 0.767 & 0.577 & 0.583  \\
\cmidrule{2-13}
 
  &\multirow{4}{*}{\rotatebox{90}{\shortstack[c]{\textbf{High}\\ \textbf{Sparsity} \\ \textbf{(99\%)}}}} & 0 & 0.752 & 0.651 & 0.578 & 0.532 & 0.108 & 0.500 & 0.487 & 0.261 & 0.289 & 0.551  \\
&& 1 & 0.812 & 0.678 & 0.643 & 0.614 & 0.135 & 0.554 & 0.527 & 0.315 & 0.327 & 0.597  \\
&& 2 & 0.889 & 0.711 & 0.728 & 0.703 & 0.175 & 0.616 & 0.598 & 0.500 & 0.428 & 0.657  \\
&& 3 & 0.919 & 0.820 & 0.800 & 0.775 & 0.234 & 0.736 & 0.647 & 0.593 & 0.539 & 0.709  \\
\cmidrule{2-13}
 
  &\multirow{4}{*}{\rotatebox{90}{\shortstack[c]{\textbf{High}\\ \textbf{Sparsity} \\ \textbf{(99.5\%)}}}} & 0 & 0.761 & 0.691 & 0.472 & 0.441 & 0.049 & 0.571 & 0.364 & 0.758 & 0.246 & 0.429  \\
&& 1 & 0.817 & 0.720 & 0.525 & 0.495 & 0.055 & 0.615 & 0.401 & 0.787 & 0.279 & 0.466  \\
&& 2 & 0.885 & 0.748 & 0.586 & 0.568 & 0.067 & 0.665 & 0.469 & 0.881 & 0.373 & 0.513  \\
&& 3 & 0.911 & 0.854 & 0.652 & 0.623 & 0.079 & 0.758 & 0.522 & 0.911 & 0.465 & 0.534  \\
\cmidrule{2-13}
 
  &\multirow{4}{*}{\rotatebox{90}{\shortstack[c]{\textbf{High}\\ \textbf{Sparsity} \\ \textbf{(99.8\%)}}}} & 0 & 0.798 & 0.770 & 0.296 & 0.387 & 0.056 & 0.618 & 0.206 & 0.824 & 0.308 & 0.622  \\
&& 1 & 0.835 & 0.787 & 0.317 & 0.411 & 0.061 & 0.647 & 0.214 & 0.835 & 0.331 & 0.647  \\
&& 2 & 0.882 & 0.807 & 0.342 & 0.439 & 0.069 & 0.678 & 0.241 & 0.893 & 0.399 & 0.677  \\
&& 3 & 0.902 & 0.863 & 0.363 & 0.469 & 0.081 & 0.741 & 0.257 & 0.929 & 0.469 & 0.699  \\

\bottomrule
\end{tabular}}
\caption{\textit{Compression Position} (see Sec.~\ref{appendix:compression}) for dense NNs (during pre-training) and $\kappa$ pruned NNs (during regularization) for all CIFAR-10 classes. Samples of a class are split into 4 subsets according to the number of human label errors in CIFAR-N to indicate the difficulty. In sparse networks, different classes are compressed at different times and difficult samples are compressed later. These results are supplemental to Tab.~3 in the main paper.}
\label{table:appendix_human_score_position}
\end{table*}

\end{document}